\documentclass[10pt,twocolumn,letterpaper]{article}

\usepackage[pagenumbers]{cvpr} 
\usepackage{booktabs,multirow}
\definecolor{cvprblue}{rgb}{0.21,0.49,0.74}
\usepackage[pagebackref,breaklinks,colorlinks,allcolors=cvprblue]{hyperref}

\usepackage[table]{xcolor}
\newsavebox{\tabtopbox}

\definecolor{bestbg}{RGB}{255,210,210}
\definecolor{secondbg}{RGB}{210,228,255}
\newcommand{\cbest}[1]{\cellcolor{bestbg}\textbf{#1}}
\newcommand{\csecond}[1]{\cellcolor{secondbg}#1}

\def\confName{CVPR}
\def\confYear{2027}

\title{Self-Calibrating Gaussian Splatting for Sparse-View Cone-Beam CT \\ under Geometric Misalignment}

\author{
Genyuan Zhang$^{1}$,
Xin Lin$^{1}$,
Wenqi Huang$^{2}$,
Zihao Wang$^{1}$,
Fenglin Liu$^{1\dagger}$,
Hongwei Bran Li$^{3}$\\[4pt]
$^{1}$Key Lab of Optoelectronic Technology and Systems, and\\
Engineering Research Center of Industrial Computed Tomography\\
Non-destructive Testing, Ministry of Education,\\
Chongqing University, Chongqing, China\\[2pt]
$^{2}$Chair for AI in Healthcare and Medicine,\\
Technical University of Munich, Munich, Germany\\[2pt]
$^{3}$Department of Diagnostic Radiology,\\
National University of Singapore, Singapore\\[4pt]
{\tt\small zhanggy@stu.cqu.edu.cn,
liufl@cqu.edu.cn}\\[2pt]
{\small 
$^{\dagger}$Corresponding authors.}
}

\begin{document}
\maketitle
\begin{abstract}
Sparse-view cone-beam computed tomography (SV-CBCT) reconstruction is highly sensitive to acquisition geometry, yet real-world acquisitions often exhibit systematic geometric misalignment. 
Importantly, we observe that the effect of geometric misalignment in Gaussian-splatting (GS) reconstruction can be largely absorbed by deforming the reconstructed volume, yielding a low projection error but an inaccurate geometry and distorted structures.
We therefore propose \textbf{CalibGS}, a self-calibrating GS framework that jointly reconstructs the volume and estimates the shared physical scanner geometry. Specifically, CalibGS constructs an identifiability-aware geometric model: residual geometry coordinates reduce coupling between geometric parameters, while object-shift compensation removes the ambiguity between geometry and Gaussian representation. Reconstruction-independent epipolar consistency is further combined with GS projection matching to provide complementary geometric constraints, and a dedicated update strategy handles the weakly identifiable geometric misalignment. 
Under large systematic misalignment, CalibGS outperforms the strongest baseline model by more than 8 dB on simulation datasets. Experiments on industrial micro-CT and laboratory CBCT systems confirm its effectiveness, and the recovered scanner-level geometry can be reused directly for full-view analytical reconstruction.
\end{abstract}    
\section{Introduction}
\label{sec:intro}
\begin{figure}[t]
  \centering           
   \includegraphics[width=\linewidth]{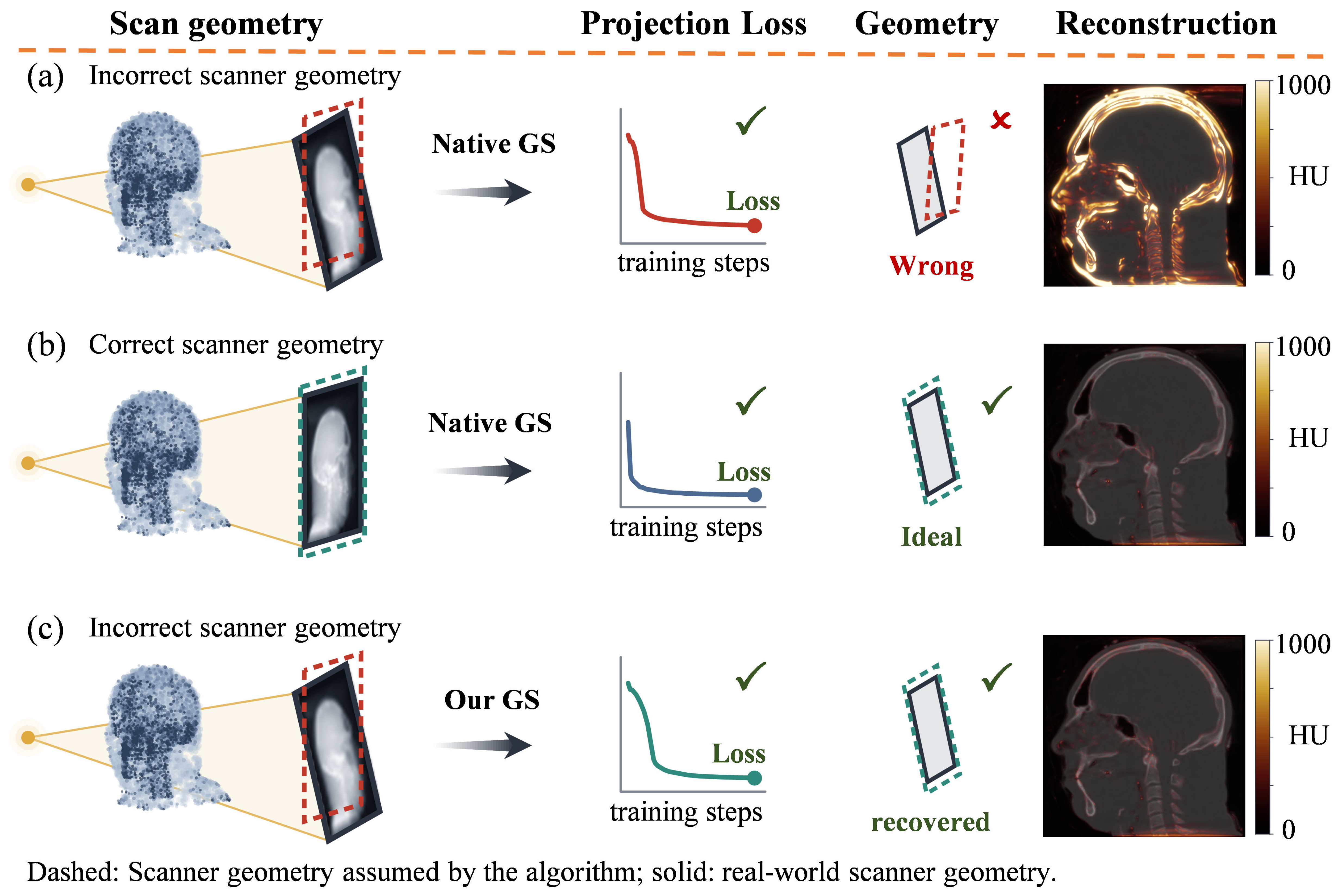}
   \caption{
    (a) Native GS optimization under an incorrect scanner geometry drives the projection loss as low as with (b) the correct scanner geometry, yet the geometry stays wrong and the reconstruction is distorted. (c) Our method uses ambiguity-aware geometry reparameterization and projection--consistency optimization for joint calibration and reconstruction.
    The last column is a sagittal slice of the reconstruction with its voxel-wise absolute reconstruction error map overlaid in color (HU).
   }
   \label{fig2}
\end{figure}

Sparse-view cone-beam computed tomography (SV-CBCT) aims to recover accurate volumetric structures from a limited number of X-ray projections~\cite{sidky2006accurate,sidky2008image}. Recent Gaussian-splatting (GS) methods have demonstrated promising reconstruction performance by representing the attenuation field with differentiable Gaussian primitives. However, these methods typically assume accurately known acquisition geometry, an assumption that is often violated in practical CBCT systems. Geometric misalignment can arise from mechanical imperfections and deviations from the nominal scanner configuration, particularly in mobile C-arm CT and industrial CBCT systems~\cite{von2004geometric,daly2008geometric,yang2006geometric,sun2006calibration}. Such errors can lead to blurred boundaries, double contours, and structural distortions~\cite{cho2005accurate}. More critically, we observe that the flexibility of GS can conceal geometric misalignment: even under incorrect scanner geometry, the Gaussian representation can adapt to reproduce the measured projections with nearly the same error as under the correct geometry, while yielding a distorted 3D reconstruction (Fig.~\ref{fig2}). This reveals a fundamental limitation of projection-driven reconstruction: \emph{accurate projection matching does not necessarily imply correct geometry or faithful volumetric recovery}.


Geometric calibration has been extensively studied in CBCT. Phantom-based methods can achieve high accuracy but require additional calibration scans and dedicated phantoms~\cite{cho2005accurate,yang2006geometric}, whereas phantom-free methods estimate geometry from projection redundancy or consistency constraints and may become less reliable when such information is limited~\cite{panetta2008optimization,aichert2015epipolar,lesaint2017calibration}. 
More recent reconstruction-based methods jointly optimize the volume and acquisition geometry~\cite{ruckert2022neat,haouchat2026geometry,choi2026pose}. However, these approaches typically assume small geometric perturbations and mainly focus on independent per-view pose errors. Sparse-view reconstruction under large, view-consistent systematic geometric misalignment therefore remains insufficiently explored.

To address this gap, a natural strategy is to jointly estimate the scanner geometry and the Gaussian representation, but this joint optimization introduces two key challenges. First, the native GS reconstruction loss is mainly based on projection matching between rendered and measured views. However, this loss does not constrain all geometric parameters equally. Flexible Gaussians can absorb part of the geometric error by transforming the reconstructed volume, so a low projection error no longer implies an accurate geometry. Second, strong coupling exists both among the geometric parameters and between the geometry and the Gaussian representation. Consequently, straightforward joint optimization of the Gaussian volume and geometry is ill-conditioned and can converge to geometrically incorrect solutions (Fig.~\ref{fig2}).

Motivated by these challenges, we propose \textbf{CalibGS}, a self-calibrating Gaussian-splatting framework for sparse-view CBCT under systematic geometric misalignment. CalibGS parameterizes the shared physical scanner geometry and embeds it into the differentiable X-ray projection process, enabling joint optimization of geometry and Gaussian representation. We first analyze the ambiguity of the physical geometry under joint GS reconstruction. The analysis reveals strong inter-parameter coupling, geometry--volume ambiguity, and particularly weak observability of the detector pitch; it also shows that GS projection matching alone provides incomplete geometric constraints. We use residual geometry parameterization to reduce inter-parameter coupling, object-shift compensation to resolve geometry--volume ambiguity, and a dedicated update strategy for the weakly identifiable detector pitch. In addition, reconstruction-independent epipolar consistency provides complementary geometric information beyond GS projection matching. Together, these designs enable stable joint calibration and sparse-view reconstruction under systematic geometric errors.

    



\begin{figure*}[t]
\centering
\includegraphics[width=\linewidth]{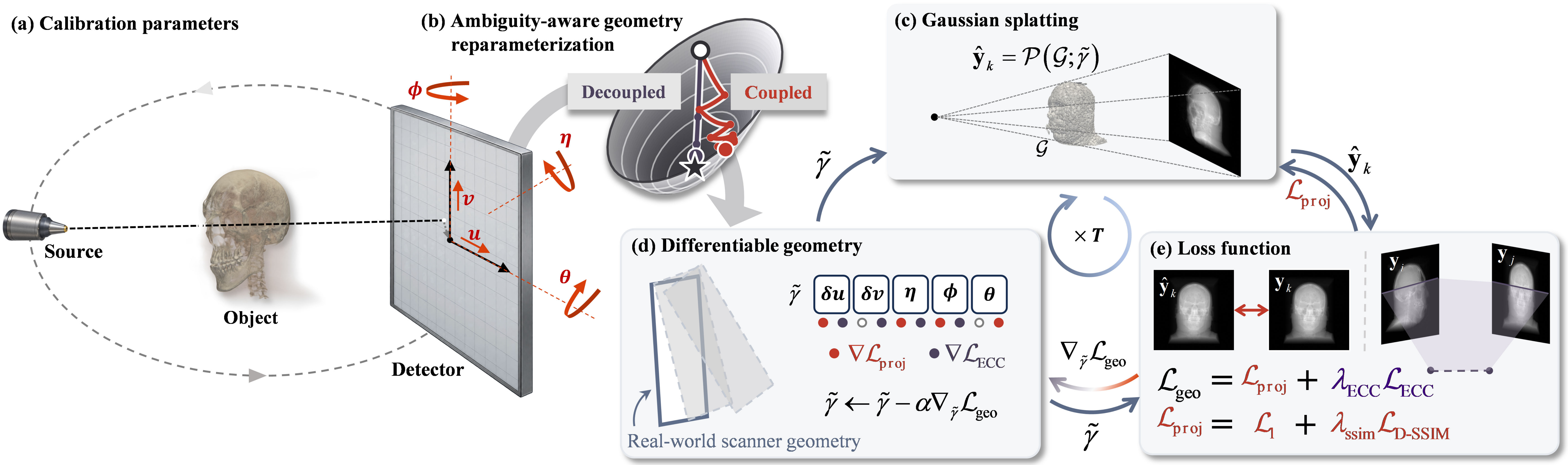}
\caption{Overview of the proposed CalibGS.
(a) CBCT acquisition geometry parameterized by the five calibration variables $(u,v,\eta,\phi,\theta)$.
(b) Ambiguity-aware geometry reparameterization for reducing coupling among geometric parameters and between geometry and the Gaussian representation.
(c) Differentiable GS reconstruction using the current geometry and Gaussian parameters.
(d) Differentiable geometry optimization guided by $\mathcal{L}_{\mathrm{geo}}$.
(e) Projection loss $\mathcal{L}_{\mathrm{proj}}$ and geometry loss $\mathcal{L}_{\mathrm{geo}}$ for joint reconstruction and calibration.}
\label{overall}
\end{figure*}

\section{Related Work}

\noindent\textbf{SV-CBCT reconstruction.}
SV-CBCT reconstruction has been extensively studied using analytical, iterative, and learning-based methods. Classical approaches include FDK~\cite{feldkamp1984practical}, iterative reconstruction~\cite{gilbert1972iterative,gordon1970art,andersen1984sart}, and sparsity-based methods such as FISTA~\cite{beck2009fista} and TV-based ASD-POCS~\cite{sidky2008image}. Deep learning further introduced learned priors through post-processing and model-based reconstruction~\cite{adler2018learned,jin2017deep,chen2018learn}, followed by diffusion-based approaches such as DiffusionMBIR~\cite{chung2023solving}. Neural fields~\cite{mildenhall2020nerf} such as NAF~\cite{zha2022naf}, NeRP~\cite{shen2022nerp}, and SAX-NeRF~\cite{cai2024structure}, and GS methods~\cite{kerbl2023gaussian} such as R$^2$-Gaussian~\cite{r2_gaussian}, 3DGR-CT~\cite{li2025threegrct}, DGR~\cite{wu2025dgr}, and FaCT-GS~\cite{pieta2026factgs}, have further improved sparse-view reconstruction. However, most existing methods assume accurately known acquisition geometry and do not explicitly address systematic scanner-level geometric misalignment.

\noindent \textbf{Geometric calibration.}
Accurate acquisition geometry is essential for CBCT reconstruction, as mechanical misalignment and system instability can cause deviations from the nominal scanner configuration. 
Conventional calibration methods estimate geometric parameters from dedicated phantoms with known structures~\cite{cho2005accurate,yang2006geometric,noo2000ellipse,mennessier2009alignment,jacobson2018line}, providing high accuracy but requiring additional calibration scans. 
Phantom-free approaches instead exploit redundancy or consistency among measured projections to estimate geometry directly from the scanned object~\cite{panetta2008optimization,aichert2015epipolar,lesaint2017calibration,luo2020misalignment}. 
More recently, differentiable reconstruction has enabled joint optimization of image representations and acquisition geometry across a range of applications, including camera parameter refinement for natural-image reconstruction~\cite{lin2021barf,jeong2021self,fu2024colmap}, motion-aware CBCT reconstruction~\cite{thies2024gradient}, self-calibrating neural fields for CT~\cite{wu2023joint,ruckert2022neat}, general differentiable tomographic calibration~\cite{haouchat2026geometry}, and splat-based pose refinement for CT~\cite{choi2026pose}.
In contrast, we parameterize the physical CBCT imaging geometry using scanner-level detector parameters shared across all views and focus on systematic geometric misalignment in sparse-view reconstruction.

\section{Method}

Figure~\ref{overall} illustrates the overall framework. The original geometric parameters in (a) are first transformed into the ambiguity-aware geometry reparameterization (b) and then passed to the differentiable geometry module (d). During optimization, the Gaussians (c) are updated by $\mathcal{L}_{\text{proj}}$, and the rendered projection $\hat{\mathbf{y}}_k$ enters the loss module (e). The geometric loss $\mathcal{L}_{\mathrm{geo}}$ is then back-propagated to update the geometric parameters in (d), and the refined geometry is fed back into (c), forming a closed-loop joint reconstruction and calibration process.

\subsection{Problem Formulation}
Given a set of SV-CBCT projections $\{\mathbf{y}_k\}_{k=1}^{K}$ acquired at known view angles $\{\beta_k\}_{k=1}^{K}$, our goal is to recover both the 3D attenuation field and the systematic geometric misalignment of the CBCT system.

Following the standard circular CBCT geometry parameterization~\cite{lesaint2017calibration}, an acquisition can be described by the source-to-isocenter distance $R$, the source-to-detector distance $D$, two detector offsets $(u,v)$ in pixels, and three detector orientation angles $(\eta,\phi,\theta)$.

In this work, we keep $R$ and $D$ fixed because they are not uniquely identifiable from the measured projections, and mainly induce global scaling rather than reconstruction artifacts. We therefore focus on the five dominant detector-misalignment parameters
\begin{equation}
    \boldsymbol{\gamma}
    =
    (u,v,\eta,\phi,\theta),
\end{equation}
where $u$ and $v$ denote the horizontal and vertical detector offsets, and $\eta$, $\phi$, and $\theta$ denote the in-plane
rotation, yaw, and pitch, respectively, as shown in Fig.~\ref{overall}. Importantly, $\boldsymbol{\gamma}$ is shared by all views and models systematic scanner-level misalignment.



Following R$^2$-Gaussian~\cite{r2_gaussian}, we represent the attenuation field using $M$ anisotropic radiative Gaussian primitives,
\begin{equation}
    \mathcal{G}
    =
    \left\{
    (\mathbf{p}_i,\mathbf{\Sigma}_i,\rho_i)
    \right\}_{i=1}^{M},
\end{equation}
where $\mathbf{p}_i\in\mathbb{R}^3$, 
$\mathbf{\Sigma}_i\in\mathbb{R}^{3\times3}$, and $\rho_i$ denote the position, covariance, and central density of the $i$-th Gaussian, respectively.
The geometry of the $k$-th view is determined by
\begin{equation}
    \Gamma_k
    =
    \Gamma(\beta_k;\boldsymbol{\gamma}),
\end{equation}
where $\boldsymbol{\gamma}$ denotes the scanner-level geometric parameters shared across all views, and $\Gamma_k$ denotes the complete acquisition geometry of view $k$.

Based on the differentiable X-ray rasterization of R$^2$-Gaussian~\cite{r2_gaussian}, the predicted projection is written as
\begin{equation}
    \hat{\mathbf{y}}_k
    =
    \mathcal{P}(\mathcal{G};\Gamma_k),
\end{equation}
where $\mathcal{P}$ denotes the differentiable X-ray forward projector.

The self-calibration problem is then formulated as jointly estimating the Gaussian representation and the shared scanner geometry:
\begin{equation}
    (\mathcal{G}^{*},\boldsymbol{\gamma}^{*})
    =
    \arg\min_{\mathcal{G},\boldsymbol{\gamma}}
    \mathcal{L}_{\mathrm{joint}}
    \left(
        \mathcal{G},
        \boldsymbol{\gamma};
        \{\mathbf{y}_k\}_{k=1}^{K}
    \right),
    \label{eq:joint_problem}
\end{equation}
where $\mathcal{L}_{\text{joint}}$ comprises $\mathcal{L}_{\text{proj}}$ (Eq.~\ref{eq:proj_loss}) for the Gaussians and $\mathcal{L}_{\text{geo}}$ (Eq.~\ref{eq:geometry_loss}) for the geometry.

\begin{figure*}[t]
  \centering
   \includegraphics[width=\linewidth]{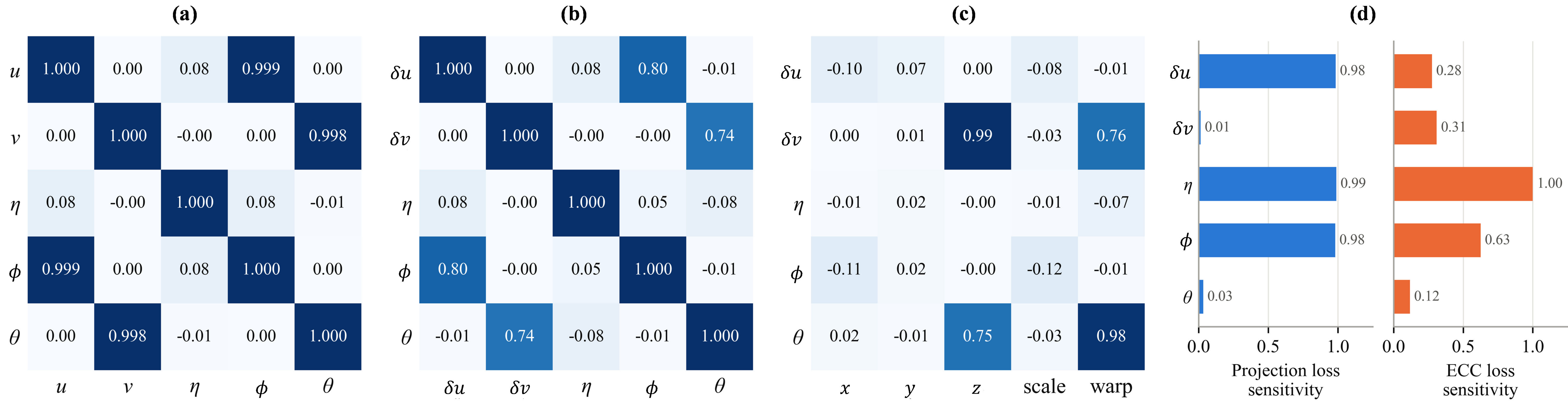}
   \caption{Correlation maps among (a) the original geometric parameters, (b) the reparameterized geometric parameters, and (c) the geometric parameters and Gaussian-volume transformations; (d) parameter-wise observability under projection matching and epipolar consistency. Blue bars indicate the fraction of projection variation that cannot be absorbed by object translation, scaling, or perspective deformation, while orange bars denote the normalized ECC sensitivity.}
   \label{fig1}
\end{figure*}

\subsection{Ambiguity-aware geometry reparameterization}
Directly optimizing the physical geometry parameters together with the Gaussian representation is poorly conditioned due to both inter-parameter coupling and geometry--volume ambiguity. We therefore first analyze the ambiguity of the CBCT geometry, as summarized in Fig.~\ref{fig1}.

\noindent \textbf{Residual geometry parameterization.}
We characterize the local effect of each geometry parameter using finite-difference projection derivatives. As shown in Fig.~\ref{fig1}(a), the raw parameters exhibit severe coupling: $u$ and $\phi$ induce nearly identical projection changes, as do $v$ and $\theta$, with correlation coefficients of 0.999 and 0.998, respectively. The resulting geometry Jacobian is highly ill-conditioned, with a condition number of approximately $2.4\times10^{5}$.

This coupling arises because detector rotations also induce apparent image-plane shifts. In particular, yaw $\phi$ contributes an effective horizontal displacement $D\tan\phi$, while pitch $\theta$ contributes an analogous vertical displacement. We therefore introduce the residual coordinates
\begin{equation}
    \delta u = u + \tfrac{D}{d_u}\tan\phi, \qquad \delta v = v + \tfrac{D}{d_v}\tan\theta,
    \label{eq:residual_geometry}
\end{equation}
where $d_u$ and $d_v$ are the detector pixel pitches, and optimize
\begin{equation}
    \tilde{\boldsymbol{\gamma}}
    =
    (\delta u,\delta v,\eta,\phi,\theta).
    \label{eq7}
\end{equation}
As shown in Fig.~\ref{fig1}(b), this reparameterization substantially reduces parameter correlation and decreases the condition number to approximately $78$. 



\noindent \textbf{Object--shift compensation.}
Figure~\ref{fig1}(c) shows that some geometry perturbations can be partially absorbed by transformations of the GS reconstructed volume. In particular, $\delta v$ is nearly equivalent to a global translation along the $z$-axis, while the effect of detector pitch $\theta$ can be approximated by a height-dependent perspective warp.

For $\delta v$, the main ambiguity arises from its near-equivalence to a global translation of the object along $z$. A translation $\Delta z$ at the isocenter produces an approximately $\frac{D}{R\,d_v}\Delta z$-pixel shift on the detector. Therefore, changes in $\delta v$ can be compensated by translations in the opposite direction of the Gaussian volume.

To remove this ambiguity, we reparameterize the Gaussian $z$-coordinates as
\begin{equation}
    z = \tilde{z} - \frac{R d_v}{D}\,\delta v,
\end{equation}
and optimize $\tilde z$ rather than $z$. Under this parameterization, changes in $\delta v$ are automatically accompanied by the corresponding object translation, making the projection loss insensitive to this ambiguous direction. As a result, $\delta v$ is estimated mainly from the reconstruction-independent consistency term introduced in Sec.~3.3.

In implementation, because the Gaussians explicitly store $z$, after each update of $\delta v$ we apply
\begin{equation}
    z \leftarrow z - \frac{R d_v}{D}\left(\delta v^{\mathrm{new}} - \delta v^{\mathrm{old}}\right),
\end{equation}
which keeps $\tilde z$ unchanged.

For the weakly identifiable pitch $\theta$, we freeze it for the first 3000 iterations, then optimize it with a $10\times$ larger step and without the ECC gradient (Secs.~3.3 and 4.4).

\subsection{Projection--Consistency Optimization}
\label{sec:collaborative}

As shown in Fig.~\ref{fig2}, projection loss alone is insufficient to calibrate all parameters, which motivates the use of a complementary consistency loss.

\noindent\textbf{Projection loss.}
For the $k$-th view, we adopt the rendering loss of
R$^2$-Gaussian~\cite{r2_gaussian},
\begin{equation}
    \mathcal{L}_{\mathrm{render}}^{(k)}
    =
    \mathcal{L}_{1}
    \left(
        \hat{\mathbf{y}}_k,\mathbf{y}_k
    \right)
    +
    \lambda_{\mathrm{ssim}}
    \mathcal{L}_{\mathrm{D\text{-}SSIM}}
    \left(
        \hat{\mathbf{y}}_k,\mathbf{y}_k
    \right).
\end{equation}
The projection objective over the $K$ measured views is
\begin{equation}
    \mathcal{L}_{\mathrm{proj}}
    =
    \frac{1}{K}
    \sum_{k=1}^{K}
    \mathcal{L}_{\mathrm{render}}^{(k)},
    \label{eq:proj_loss}
\end{equation}

\begin{figure*}[t]
  \centering
   \includegraphics[width=\linewidth]{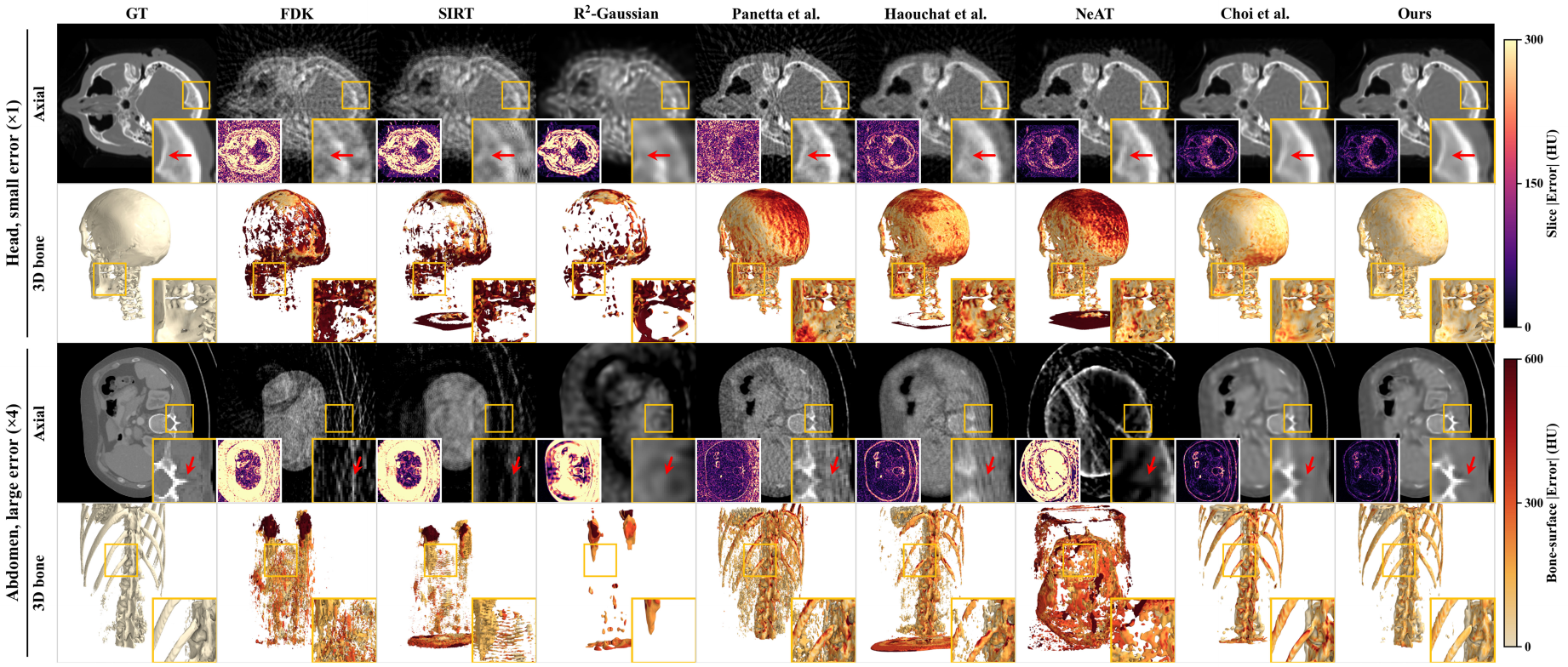}
   \caption{Qualitative comparison on simulated data (50 views).
    Top two rows: head ($\times$1); bottom two rows: abdomen ($\times$4). Axial: bottom left, $|$recon$-$GT$|$ (0--300\,HU); bottom right, zoom of the yellow box. 3D: bone surface colored by $|$recon$-$GT$|$ (0--600\,HU). Red arrows indicate regions of interest.}
   \label{x1}
\end{figure*}

\begin{figure*}[t]
  \centering
   \includegraphics[width=\linewidth]{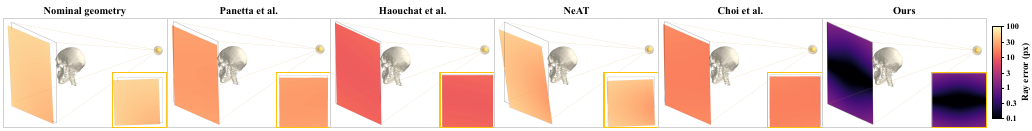}
   \caption{Estimated scanner geometry (head, $\times$1 error). Black frame: real detector; colored plate:
detector placed by each method's estimated geometry; inset: view from the
source. Color: ray error, \ie the distance between where each pixel's ray is assumed to land and where it truly
lands (px, log scale). }
   \label{geometry}
\end{figure*}

\noindent\textbf{Epipolar consistency loss.}
By the epipolar consistency conditions (ECC) based on the Grangeat's relation~\cite{grangeat1991framework, aichert2015epipolar}, two measured projections $i$ and $j$ should provide the same derivative of the 3D Radon transform on an epipolar plane $E$ passing through both source positions. We therefore define
\begin{equation}
    \mathcal{L}_{\mathrm{ECC}}
    =
    \frac{1}{|\Omega|}
    \sum_{(i,j)\in\Omega}
    \frac{1}{|\mathcal{E}_{ij}|}
    \sum_{E\in\mathcal{E}_{ij}}
    \left\|
        \mathcal{C}_{i}(E;\tilde{\boldsymbol{\gamma}})
        -
        \mathcal{C}_{j}(E;\tilde{\boldsymbol{\gamma}})
    \right\|_2^2,
    \label{eq:ecc_loss}
\end{equation}
where $\Omega$ denotes the set of sampled projection pairs, $\mathcal{E}_{ij}$ denotes the set of epipolar planes associated with pair $(i,j)$, and $\mathcal{C}_{i}$ is the corresponding Grangeat-domain quantity computed from the measured projection (see supplementary material for the explicit Grangeat formulation and differentiable implementation).

\noindent\textbf{Collaborative geometry optimization.}
The two losses provide complementary geometric information, as shown in Fig.~\ref{fig1}(d). Projection matching is strongly sensitive to $\delta u$, $\eta$, and $\phi$, but is nearly blind to $\delta v$. In contrast, epipolar consistency provides strong constraints on $\delta u$, $\delta v$, $\eta$, and $\phi$, but contains little reliable information about $\theta$. We therefore combine them for geometry optimization as
\begin{equation}
    \mathcal{L}_{\mathrm{geo}}
    =
    \mathcal{L}_{\mathrm{proj}}
    +
    \lambda_{\mathrm{ECC}}
    \mathcal{L}_{\mathrm{ECC}}.
    \label{eq:geometry_loss}
\end{equation}
Accordingly, $\delta u$, $\eta$, and $\phi$ are constrained by both, and $\delta v$ mainly by the epipolar consistency. $\theta$ is weakly constrained by both: the ECC cost varies with $\theta$ but its minimum is biased (Fig.~\ref{fig:theta_strategy}), so $\theta$ is excluded from $\mathcal{L}_{\text{ECC}}$ and estimated by projection matching alone.
Finally, as an optional refinement, we fix the calibrated geometry to perform full-capacity Gaussian reconstruction.

\newsavebox\topbox\newlength\mw
\begin{table*}[t]\centering \caption{Quantitative comparison on simulated data (50 views). Top: registered 3D PSNR (dB)/SSIM, downstream segmentation (mean Dice over the 30 head and 40 abdominal structures) and run time. Bottom: absolute errors of the estimated scanner geometry. $^\dagger$Not estimated (nominal value kept). $^\ddagger$Per-view poses: the global geometry that best reproduces the mean learned camera correction. \colorbox{bestbg}{\textbf{Best}} and \colorbox{secondbg}{second best} results are highlighted.} \label{tab:main} \setlength{\tabcolsep}{3.5pt}\setlength{\aboverulesep}{0pt}\setlength{\belowrulesep}{0pt}\renewcommand{\arraystretch}{1.12}\small
\settowidth\mw{Ours (w/o refinement)}
\sbox\topbox{\begin{tabular}{l cccccccc cccc c}
\toprule
\multirow{2}{*}{Method} & \multicolumn{4}{c}{Head (PSNR/SSIM)} & \multicolumn{4}{c}{Abdomen (PSNR/SSIM)} & \multicolumn{2}{c}{Head (Dice)} & \multicolumn{2}{c}{Abd. (Dice)} & \multirow{2}{*}{Time} \\
\cmidrule(lr){2-5}\cmidrule(lr){6-9}\cmidrule(lr){10-11}\cmidrule(lr){12-13}
  & \multicolumn{2}{c}{$\times1$} & \multicolumn{2}{c}{$\times4$} & \multicolumn{2}{c}{$\times1$} & \multicolumn{2}{c}{$\times4$} & $\times1$ & $\times4$ & $\times1$ & $\times4$ & \\
\midrule
FDK~[14] & \multicolumn{2}{c}{23.69/0.375} & \multicolumn{2}{c}{19.14/0.314} & \multicolumn{2}{c}{21.07/0.259} & \multicolumn{2}{c}{17.84/0.186} & 0.202 & 0.014 & 0.069 & 0.000 & 2\,s \\
SIRT~[15] & \multicolumn{2}{c}{24.70/0.648} & \multicolumn{2}{c}{18.80/0.396} & \multicolumn{2}{c}{22.14/0.591} & \multicolumn{2}{c}{17.86/0.386} & 0.171 & 0.012 & 0.059 & 0.000 & 23\,min \\
R$^2$-Gaussian~[44] & \multicolumn{2}{c}{25.44/0.778} & \multicolumn{2}{c}{18.63/0.549} & \multicolumn{2}{c}{23.72/0.738} & \multicolumn{2}{c}{18.68/0.596} & 0.189 & 0.016 & 0.064 & 0.001 & 7\,min \\
\midrule
Panetta \etal~[28] & \multicolumn{2}{c}{28.34/0.488} & \multicolumn{2}{c}{25.57/0.486} & \multicolumn{2}{c}{25.21/0.376} & \multicolumn{2}{c}{23.23/0.373} & 0.713 & 0.683 & 0.789 & 0.748 & 3\,s \\
Haouchat \etal~[18] & \multicolumn{2}{c}{31.42/0.707} & \multicolumn{2}{c}{27.05/0.658} & \multicolumn{2}{c}{28.19/0.571} & \multicolumn{2}{c}{26.60/0.546} & 0.701 & 0.635 & 0.726 & 0.598 & 91\,min \\
NeAT~[30] & \multicolumn{2}{c}{30.39/0.867} & \multicolumn{2}{c}{17.20/0.398} & \multicolumn{2}{c}{26.19/0.780} & \multicolumn{2}{c}{17.13/0.488} & 0.740 & 0.004 & 0.533 & 0.000 & 36\,min \\
Choi \etal~[10] & \multicolumn{2}{c}{36.71/0.966} & \multicolumn{2}{c}{30.25/0.935} & \multicolumn{2}{c}{32.13/0.930} & \multicolumn{2}{c}{27.67/0.892} & \csecond{0.879} & 0.816 & 0.837 & 0.769 & 7\,min \\
Ours (w/o refinement) & \multicolumn{2}{c}{\csecond{39.00/0.973}} & \multicolumn{2}{c}{\csecond{37.95/0.967}} & \multicolumn{2}{c}{\csecond{36.27/0.951}} & \multicolumn{2}{c}{\csecond{34.81/0.941}} & \csecond{0.879} & \csecond{0.870} & \csecond{0.879} & \csecond{0.853} & 12\,min \\
Ours & \multicolumn{2}{c}{\cbest{39.70/0.976}} & \multicolumn{2}{c}{\cbest{39.49/0.975}} & \multicolumn{2}{c}{\cbest{36.65/0.953}} & \multicolumn{2}{c}{\cbest{35.80/0.951}} & \cbest{0.892} & \cbest{0.906} & \cbest{0.893} & \cbest{0.889} & 20\,min \\
\midrule[\heavyrulewidth]\end{tabular}}
\usebox\topbox\par\nointerlineskip
\begin{tabular*}{\wd\topbox}{@{\extracolsep{\fill}\hspace{\tabcolsep}}p{\mw} *{10}{c}}
\multirow{2}{*}{Method} & \multicolumn{5}{c}{Geometry error, $\times1$} & \multicolumn{5}{c}{Geometry error, $\times4$} \\ \cmidrule(lr){2-6}\cmidrule(lr){7-11}
  & $|\Delta u|$ & $|\Delta v|$ & $|\eta|$ & $|\phi|$ & $|\theta|$ & $|\Delta u|$ & $|\Delta v|$ & $|\eta|$ & $|\phi|$ & $|\theta|$ \\ \midrule
Panetta \etal~[28] & 0.107 & 5.82$^\dagger$ & 0.092 & 0.339 & 0.50$^\dagger$ & \csecond{0.091} & 23.29$^\dagger$ & 0.098 & \csecond{0.427} & 2.00$^\dagger$ \\
Haouchat \etal~[18] & \csecond{0.016} & \csecond{3.99} & 0.007 & 0.103 & \csecond{0.39} & 0.112 & \csecond{7.06} & \csecond{0.016} & 0.451 & \csecond{1.22} \\
NeAT$^\ddagger$~[30] & 0.359 & 4.60 & 0.434 & 0.337 & 4.75 & 2.146 & 31.35 & 2.068 & 0.652 & 8.16 \\
Choi \etal$^\ddagger$~[10] & 0.046 & 4.20 & \csecond{0.006} & \csecond{0.035} & 0.40 & 0.579 & 14.02 & 0.050 & 0.457 & 1.38 \\
Ours & \cbest{0.003} & \cbest{0.09} & \cbest{0.000} & \cbest{0.003} & \cbest{0.21} & \cbest{0.007} & \cbest{0.12} & \cbest{0.001} & \cbest{0.004} & \cbest{0.71} \\
\bottomrule\end{tabular*}\end{table*}\null

\section{Experiments}

\subsection{Experimental Setup}
\noindent\textbf{Simulated datasets.}
We used the head and abdomen CT datasets ($256^3$) from the R$^2$-Gaussian benchmark~\cite{r2_gaussian}.
We simulated 50 views over $360^\circ$ for training and 100 for testing with TIGRE~[6], adding Poisson ($I_0{=}10^5$) and  Gaussian noise ($\sigma=10$ photon counts).
The detector was misaligned by $[u,v,\eta,\phi,\theta]=s\cdot[2\,\text{px},-2\,\text{px},0.5^\circ,0.5^\circ,0.5^\circ]$ with $s{=}1$ (small) or $s{=}4$ (large). The supplementary materials provide additional experiments covering diverse phantoms and geometric misalignments.

\begin{figure}[t]
  \centering
   \includegraphics[width=0.9\linewidth]{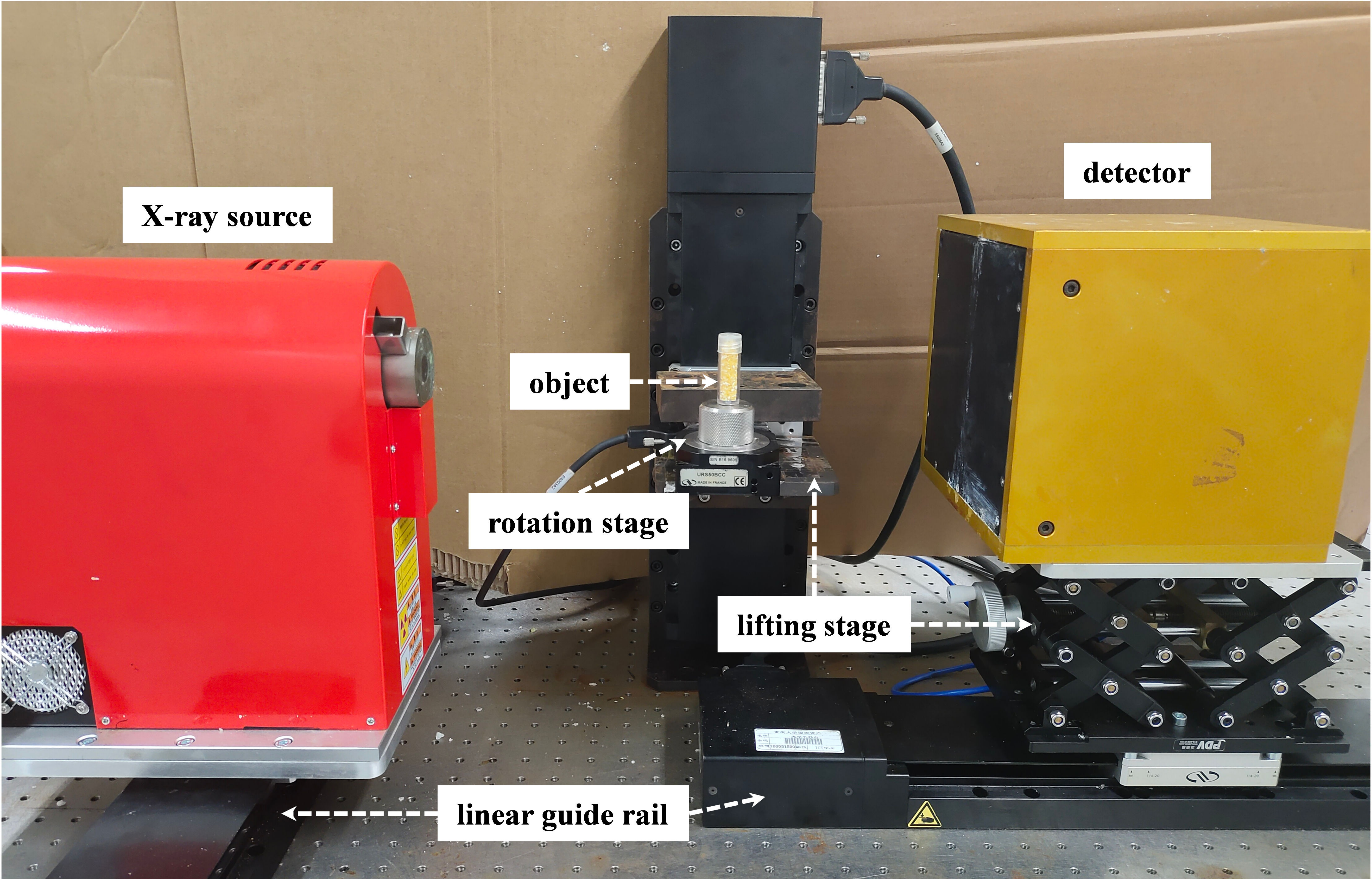}
   \caption{Our laboratory CBCT system (Corn and Beans).}
   \label{X-ray}
\end{figure}

\noindent\textbf{Real-world acquisition datasets.}
We evaluated CalibGS on the CWI walnut dataset~\cite{der2019cone} and a corn-kernel dataset acquired using our laboratory CBCT system. The acquisition system is shown in Fig. \ref{X-ray}. Each method used 50 views for calibration and 100 held-out views for evaluation. For Walnut, the 1200-view FDK~\cite{feldkamp1984practical} reconstruction using the CWI calibration served as the 3D reference. Additional industrial and laboratory scans were used to evaluate transferability of the estimated geometry to full-view FDK reconstruction.

\noindent\textbf{Metrics.}
We reported registered 3D PSNR/SSIM~\cite{avants2011reproducible, wang2009mean, wang2004image}, Dice score in downstream image segmentation task, and absolute geometric errors on simulated data. For real data, geometry accuracy was additionally evaluated using 100 held-out measured projections that were excluded from optimization~\cite{ruckert2022neat}. Run time was also reported.

\noindent\textbf{Implementation and baselines.}
CalibGS was built upon R$^2$-Gaussian~\cite{r2_gaussian} and followed the joint calibration and refinement stage described in Sec.~3. We compared our approach against traditional reconstruction~\cite{feldkamp1984practical,gilbert1972iterative}, traditional geometric calibration~\cite{panetta2008optimization}, sparse-view reconstruction~\cite{r2_gaussian}, and joint calibration and reconstruction~\cite{haouchat2026geometry,ruckert2022neat,choi2026pose}. Detailed acquisition parameters, evaluation protocols, implementation settings, and baseline adaptations are provided in the supplementary material.

\subsection{Results on Simulated Datasets}
\label{sec:sim_results}

Figure~\ref{x1} and Table~\ref{tab:main} compare the methods.
Under the nominal geometry, FDK, SIRT and R$^2$-Gaussian show blur, doubled contours and streaks, which become more severe at $\times4$.
Panetta \etal correct only $u,\eta,\phi$, so the uncorrected $v$ and $\theta$ leave residual blur on top of sparse-view FDK streaks.
Haouchat \etal estimate all five parameters, but their smooth B-spline volume cannot fit fine details, leaving edges blurred.
NeAT handles small errors but diverges at $\times4$.
The method of Choi \etal\ is the strongest baseline, but systematic geometric errors are partially absorbed by both the per-view poses and the reconstructed volume, making optimization difficult under large misalignment. Its PSNR drops by 4.5-–6.5\,dB from $\times1$ to $\times4$, with distorted bone boundaries and structured residuals along these edges (zoomed regions).
Registration exposes such residual errors: it raises the PSNR of Choi \etal by up to 11.4\,dB, as its poses drift together with the object, but changes ours by at most 0.18\,dB. In contrast, our single five-parameter geometry is shared by all views, the ECC term constrains it directly from the measured projections, and object-shift compensation mitigates the coupling between $\delta v$ and an object shift.
Figure~\ref{geometry} and Table~\ref{tab:main} show that all parameters except the weakly identifiable $\theta$ are estimated with sub-pixel and sub--0.01$^\circ$ accuracy.
Our method loses at most 0.85\,dB from $\times1$ to $\times4$, outperforms the best baseline at $\times4$ by 9.2/8.1\,dB (head/abdomen), and yields sharp bone edges and the smallest errors at both levels.
The gains carry over to downstream segmentation: ours attains the highest Dice in all four settings and loses at most 0.004 from $\times$1 to $\times$4, whereas Choi et al.\ lose 0.06--0.07.
Ours (w/o refinement) alone already surpasses all baselines in reconstruction PSNR and Dice, and the full pipeline (20\,min) is faster than Haouchat \etal (91\,min) and NeAT (36\,min).

\begin{figure*}[t]
  \centering
   \includegraphics[width=\linewidth]{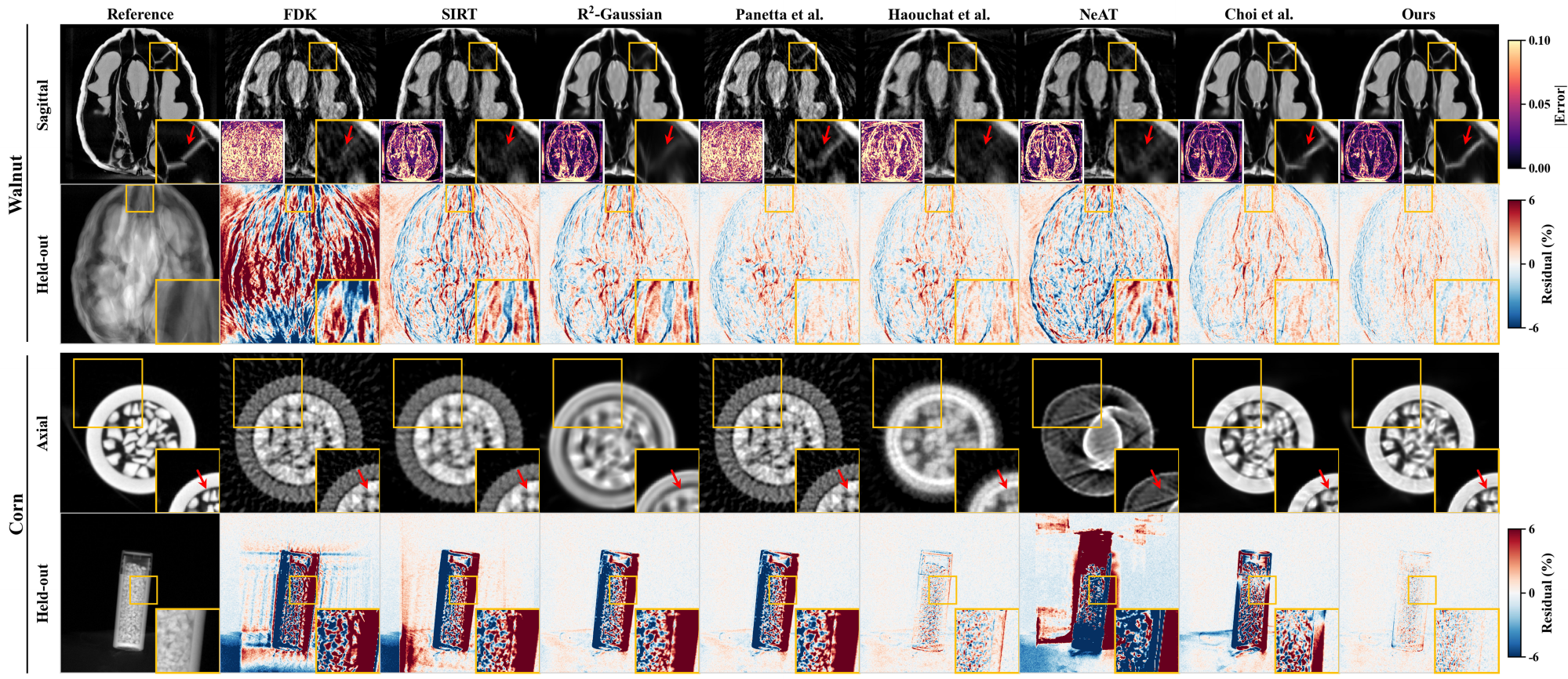}
   \caption{Real-world Acquisition Datasets reconstruction comparison on Walnut and Corn (50 views). Top: Walnut sagittal reconstructions and held-out projection residuals. Bottom: Corn axial reconstructions and held-out projection residuals. Red arrows indicate regions of interest, and insets show enlarged views.}
   \label{real}
\end{figure*}

\begin{table}[t]
\centering
\caption{Quantitative results on real data (50 views).
3D: registered PSNR (dB)/SSIM.
Held-out: PSNR/SSIM of 100 measured projections.
$^\ddagger$R$^2$-Gaussian trained with the method's geometry.
$^\S$Mean training-view pose.
$^\ast$R$^2$-Gaussian with CWI's calibrated geometry fixed, trained with the exact same fixed-geometry reconstruction setting as our refinement stage.}
\label{tab:real}
\setlength{\tabcolsep}{4pt}
\setlength{\aboverulesep}{0pt}
\setlength{\belowrulesep}{0pt}
\renewcommand{\arraystretch}{1.12}
\footnotesize
\begin{tabular}{l cc c}
\toprule
\multirow{2}{*}{Method} & \multicolumn{2}{c}{Walnut} & Corn \\
\cmidrule(lr){2-3}\cmidrule(lr){4-4}
& 3D & Held-out & Held-out \\
\midrule
FDK~\cite{feldkamp1984practical}             & 17.91/0.227 & -- & -- \\
SIRT~\cite{gilbert1972iterative}             & 22.88/0.483 & -- & -- \\
R$^2$-Gaussian~\cite{r2_gaussian}            & 24.20/0.564 & 35.12/0.916 & 24.10/0.880 \\
Panetta \etal~\cite{panetta2008optimization} & 17.98/0.235 & \csecond{36.87/0.929}$^\ddagger$ & 24.09/0.881$^\ddagger$ \\
Haouchat \etal~\cite{haouchat2026geometry}   & 20.65/0.296 & 36.44/0.925$^\ddagger$ & \csecond{38.91/0.929}$^\ddagger$ \\
NeAT~\cite{ruckert2022neat}                  & 20.57/0.476 & 31.93/0.863$^\S$ & 18.19/0.593$^\S$ \\
Choi \etal~\cite{choi2026pose}               & \csecond{24.79/0.579} & 36.05/0.925$^\S$ &  27.90/0.866$^\S$ \\
Ours                                         & \cbest{25.05/0.586} & \cbest{38.39/0.936} & \cbest{40.79/0.933} \\
\midrule
\textit{CWI geometry}$^\ast$                 & 25.14/0.589 & 37.47/0.932 & n/a \\
\bottomrule
\end{tabular}
\end{table}

\subsection{Results on Real-World Acquisition Datasets}
\label{sec:real_results}

Figure~\ref{real} and Table~\ref{tab:real} present the results on Walnut and Corn. On Walnut, FDK, SIRT, and Panetta \etal\ exhibit streaks or doubled structures, while R$^2$-Gaussian, Haouchat \etal, and NeAT produce blurred reconstructions. Choi \etal\ perform better but still exhibit structural distortions in the enlarged region. Our method achieves the highest 3D and held-out-view PSNR. Notably, it outperforms the CWI calibration by 0.92\,dB in held-out-view PSNR under the same reconstruction pipeline, indicating that the improvement stems from more accurate geometry calibration. Our estimated geometry closely agrees with the CWI calibration, except for a 4.2\,px difference in $v$. Since errors in $v$ can be largely absorbed by vertical translations of the reconstructed volume (Sec.~3.2), such discrepancies may be difficult to detect through reconstruction-based calibration alone.

On Corn, where no ground-truth volume is available, the 1024-view FDK reconstruction using our estimated geometry serves only as a visual reference. The laboratory scanner exhibits substantially larger geometric deviations than the Walnut scanner. Methods using the nominal geometry and Panetta \etal\ produce a doubled bottle wall, Haouchat \etal\ recover the axis offset but blur the kernels, and NeAT fails under large geometric errors. Compared with Choi \etal, our method produces clearer structures and substantially better quantitative results, recovering a sharp bottle wall and well-separated kernels while achieving the highest held-out-view PSNR.




\begin{table}[t]
\centering
\caption{Ablation study of Ours (w/o refinement) on the abdomen dataset ($\times4$, 50 views).}
\label{tab:ablation}
\setlength{\tabcolsep}{2.6pt}
\setlength{\aboverulesep}{0pt}
\setlength{\belowrulesep}{0pt}
\renewcommand{\arraystretch}{1.12}
\footnotesize
\begin{tabular}{l ccccc cc}
\toprule
Setting & $|\Delta u|$ & $|\Delta v|$ & $|\eta|$ & $|\phi|$ & $|\theta|$ & PSNR & SSIM \\
\midrule
Ours (w/o refinement)         & 0.012 & 0.16  & 0.001 & 0.000 & 0.69 & \cbest{34.81}   & \cbest{0.941} \\
\midrule
w/o ECC              & 0.012 & 18.95 & 0.001 & 0.011 & 4.91 & 25.74 & 0.883 \\
w/o object-shift comp.      & 0.011 & 2.33  & 0.001 & 0.001 & 4.46 & 29.53 & 0.907 \\
w/o residual param.  & 0.102 & 2.61  & 0.011 & 0.296 & 2.57 & 31.25 & 0.921 \\
\bottomrule
\end{tabular}
\end{table}

\subsection{Ablation Study}
\label{sec:ablation}

\noindent\textbf{Components.}
Table~\ref{tab:ablation} shows that without ECC, $\Delta v$ is unconstrained, which drift by 19\,px and dragged $\theta$ along, costing 9.1\,dB.
Without object-shift compensation, the Gaussians compensated $\Delta v$ by an object shift and $\theta$ drifted to $4.5^\circ$ at $\times4$ ($-5.3$\,dB).
Without residual geometry parameterization, $u$--$\phi$ and $v$--$\theta$ remain strongly coupled, raising the $\phi$ error to $0.3^\circ$ and the $\theta$ error to $2.6^\circ$ ($-3.6$\,dB).


\begin{figure}[t]
\centering
\includegraphics[width=\linewidth]{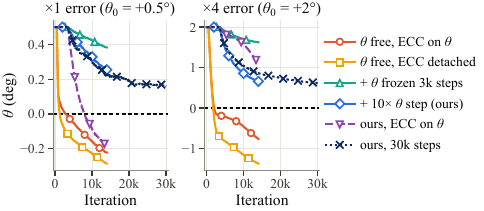}
\caption{Optimization trajectories of the weakly identifiable pitch parameter $\theta$ under different strategies.}
\label{fig:theta_strategy}
\end{figure}

\noindent \textbf{Pitch optimization.}
Figure~\ref{fig:theta_strategy} evaluates the optimization strategy for the weakly identifiable pitch $\theta$ under both small ($\theta_0=0.5^\circ$) and large ($\theta_0=2^\circ$) errors. Directly optimizing $\theta$ from the beginning leads to unstable updates, while detaching the ECC gradient alone is insufficient. Freezing $\theta$ during the early stage stabilizes the trajectory, and the enlarged $10\times$ angular step further improves its correction, forming our final strategy. Re-enabling ECC on $\theta$ introduces noticeable bias, confirming that ECC provides unreliable guidance along this direction. Extending the optimization to 30k iterations yields only marginal improvement, indicating that the remaining error mainly arises from weak identifiability rather than insufficient convergence.

\section{Discussion}
\noindent\textbf{Robustness to low dose and ultra sparse views.}
As shown in Table~\ref{tab:extreme}, CalibGS maintains sub-pixel translational accuracy ($|\Delta u|, |\Delta v| < 1$\,px) and sub-$0.05^\circ$ orientation accuracy in $\eta$ and $\phi$ even when the photon count drops by $50\times$ ($I_0=2\times10^3$) or the view count is reduced to 10 under large misalignment ($\times 4$).
This resilience stems from two properties: (1) optimizing a shared scanner geometry rather than per-view poses prevents pose drift under extreme view sparsity, with global Gaussian rendering and epipolar consistency jointly constraining all parameters; and (2) noise primarily affects the weakly identifiable $\delta v$ and $\theta$, consistent with Fig.~\ref{fig1}(d).
Consequently, our method is well-suited for phantom-free calibration even when only low-dose or few-view scans are available.

\begin{table}[t]
\centering
\caption{Our method under low dose and ultra sparse views (abdomen, $\times4$ error).
Absolute geometric errors and registered 3D PSNR (dB)/SSIM of the final reconstruction.
Standard: 50 views over $360^\circ$, $I_0{=}10^5$.}
\label{tab:extreme}
\setlength{\tabcolsep}{2.6pt}
\setlength{\aboverulesep}{0pt}
\setlength{\belowrulesep}{0pt}
\renewcommand{\arraystretch}{1.12}
\footnotesize
\begin{tabular}{l ccccc cc}
\toprule
Condition & $|\Delta u|$ & $|\Delta v|$ & $|\eta|$ & $|\phi|$ & $|\theta|$ & PSNR & SSIM \\
\midrule
Standard                    & 0.012 & 0.16 & 0.001 & 0.000 & 0.69 & 35.80 & 0.951 \\
\midrule
$I_0{=}10^4$                & 0.017 & 0.43 & 0.002 & 0.001 & 0.62 & 35.16 & 0.936 \\
$I_0{=}2{\times}10^3$       & 0.021 & 0.90 & 0.000 & 0.004 & 1.09 & 32.18 & 0.872 \\
\midrule
20 views                    & 0.002 & 0.14 & 0.001 & 0.006 & 0.38 & 30.04 & 0.876 \\
10 views                    & 0.008 & 0.05 & 0.000 & 0.043 & 0.75 & 27.07 & 0.811 \\
\bottomrule
\end{tabular}
\end{table}

\begin{figure}[t]
  \centering
   \includegraphics[width=\linewidth]{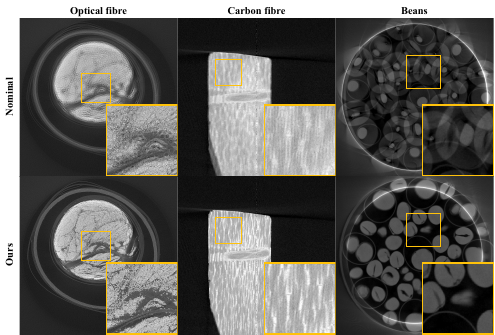}
   \caption{FDK reconstructions from all views of three additional real scans, under the nominal geometry and under our geometry calibrated from 50 views: optical fibre, carbon-fibre composite and beans. Bottom right: zoom of the yellow box.}
   \label{realfdk}
\end{figure}

\noindent\textbf{Reusable physical geometry calibration.}
Unlike per-view poses, our five parameters describe the scanner itself and can be reused by any reconstruction algorithm and for views not used in the calibration.
We estimated the geometry from 50 views and then applied it to full-view FDK reconstruction for three additional scans: an optical-fibre bundle and a carbon-fibre composite acquired on an industrial micro-CT system, and beans acquired on our laboratory CBCT.
With the nominal geometry, the full-view FDK was blurred and show doubled edges (Fig.~\ref{realfdk}).
With our geometry, the individual fibres of the bundle are resolved, the fibre tows of the composite became sharp, and the beans show clear boundaries and internal cracks.
Thus, a calibration from a sparse subset suffices to correct the full scan, making our method a practical tool for conventional FDK pipelines as well.

\noindent\textbf{Limitations.}
First, $R$ and $D$ are taken as known.
Errors in these distances change the magnification, which the five parameters cannot represent.
Second, $\theta$ remains the least observable parameter.
Our $\theta$ strategy reduces its error but does not eliminate it: 35--42\,\% of the initial error remains (Tab.~\ref{tab:main}), because the Gaussians partly absorb the residual tilt as a perspective warp.
Third, our method is slower than R$^2$-Gaussian and Choi \etal, mainly because of the ECC term, which evaluates many projection pairs per step.

\section{Conclusion}
We proposed CalibGS, a self-calibrating GS framework for sparse-view CBCT under geometric misalignment. By explicitly modeling the shared physical scanner geometry, CalibGS jointly reconstructs the volume and estimates the acquisition geometry. Ambiguity-aware geometry reparameterization and reconstruction-independent ECC loss provide complementary geometric constraints beyond projection matching alone. Experiments on simulated and real-world acquired scans, including low-dose and 10-view settings, demonstrate accurate and robust calibration, while the estimated scanner geometry can be directly reused by other reconstruction methods.

\section*{LLM Usage Statement}
ChatGPT (OpenAI) was used to assist with English language polishing, grammatical correction, and improving the clarity of manuscript presentation. Claude Opus 5.5 (Anthropic) was used, under author supervision, to assist with experimental data organization, preprocessing scripts, and routine code refactoring and debugging. In addition, ChatGPT (OpenAI) was used to generate selected illustrative elements in the schematic figures; these elements were used only for visual presentation and did not represent experimental data or results. All AI-assisted text, code, data-processing procedures, and visual elements were manually inspected, verified, and revised by the authors when necessary. The research questions, methodological design, experimental protocols, analysis and interpretation of results, and scientific conclusions were determined by the authors, who take full responsibility for all content in this submission.
{
    \small
    \bibliographystyle{ieeenat_fullname}
    \bibliography{main}

@String(ICCV= {Int. Conf. Comput. Vis.})

@String(ICCV  = {ICCV})

@article{sidky2006accurate,
  title={Accurate image reconstruction from few-views and limited-angle data in divergent-beam {CT}},
  author={Sidky, Emil Y and Kao, Chien-Min and Pan, Xiaochuan},
  journal={Journal of X-ray Science and Technology},
  volume={14},
  number={2},
  pages={119--139},
  year={2006},
  publisher={SAGE Publications Sage UK: London, England}
}

@article{sidky2008image,
  title={Image reconstruction in circular cone-beam computed tomography by constrained, total-variation minimization},
  author={Sidky, Emil Y and Pan, Xiaochuan},
  journal={Physics in Medicine \& Biology},
  volume={53},
  number={17},
  pages={4777--4807},
  year={2008}
}

@inproceedings{r2_gaussian,
  title     = {{R$^2$-Gaussian}: Rectifying Radiative Gaussian Splatting for Tomographic Reconstruction},
  author    = {Ruyi Zha and Tao Jun Lin and Yuanhao Cai and Jiwen Cao and Yanhao Zhang and Hongdong Li},
  booktitle = {Advances in Neural Information Processing Systems (NeurIPS)},
  year      = {2024}
}

@article{daly2008geometric,
  title={Geometric calibration of a mobile {C}-arm for intraoperative cone-beam {CT}},
  author={Daly, MJ and Siewerdsen, JH and Cho, YB and Jaffray, DA and Irish, JC},
  journal={Medical Physics},
  volume={35},
  number={5},
  pages={2124--2136},
  year={2008},
  publisher={Wiley Online Library}
}

@article{von2004geometric,
  title={Geometric misalignment and calibration in cone-beam tomography},
  author={Von Smekal, Lorenz and Kachelrie{\ss}, Marc and Stepina, Elizaveta and Kalender, Willi A},
  journal={Medical Physics},
  volume={31},
  number={12},
  pages={3242--3266},
  year={2004},
  publisher={Wiley Online Library}
}

@article{panetta2008optimization,
  title={An optimization-based method for geometrical calibration in cone-beam {CT} without dedicated phantoms},
  author={Panetta, D and Belcari, Nicola and Del Guerra, Alberto and Moehrs, S},
  journal={Physics in Medicine \& Biology},
  volume={53},
  number={14},
  pages={3841--3861},
  year={2008}
}

@article{lesaint2017calibration,
  title={Calibration for circular cone-beam {CT} based on consistency conditions},
  author={Lesaint, J{\'e}r{\^o}me and Rit, Simon and Clackdoyle, Rolf and Desbat, Laurent},
  journal={IEEE Transactions on Radiation and Plasma Medical Sciences},
  volume={1},
  number={6},
  pages={517--526},
  year={2017},
  publisher={IEEE}
}

@article{sun2006calibration,
  title={A calibration method for misaligned scanner geometry in cone-beam computed tomography},
  author={Sun, Yi and Hou, Ying and Zhao, Fengyong and Hu, Jiasheng},
  journal={NDT \& E International},
  volume={39},
  number={6},
  pages={499--513},
  year={2006},
  publisher={Elsevier}
}

@article{yang2006geometric,
  title={A geometric calibration method for cone beam {CT} systems},
  author={Yang, Kai and Kwan, Alexander LC and Miller, DeWitt F and Boone, John M},
  journal={Medical Physics},
  volume={33},
  number={6Part1},
  pages={1695--1706},
  year={2006},
  publisher={Wiley Online Library}
}

@article{cho2005accurate,
  title={Accurate technique for complete geometric calibration of cone-beam computed tomography systems},
  author={Cho, Youngbin and Moseley, Douglas J and Siewerdsen, Jeffrey H and Jaffray, David A},
  journal={Medical Physics},
  volume={32},
  number={4},
  pages={968--983},
  year={2005},
  publisher={Wiley Online Library}
}

@article{ruckert2022neat,
  title   = {{NeAT}: Neural Adaptive Tomography},
  author  = {R{\"u}ckert, Darius and Wang, Yuanhao and Li, Rui and Idoughi, Ramzi and Heidrich, Wolfgang},
  journal = {ACM Transactions on Graphics},
  volume  = {41},
  number  = {4},
  pages   = {55:1--55:13},
  year    = {2022},
  doi     = {10.1145/3528223.3530121}
}

@article{haouchat2026geometry,
  title   = {Geometry Calibration in Tomography with a Differentiable Ray-Based Model},
  author  = {Haouchat, Youssef and Boquet-Pujadas, Aleix and Kashani, Sepand and Th{\'e}venaz, Philippe and Unser, Michael},
  journal = {arXiv preprint arXiv:2606.21405},
  year    = {2026}
}

@inproceedings{choi2026pose,
  title     = {Revisiting Pose Sensitivity in Splat-based Computed Tomography under Sparse-view Reconstruction},
  author    = {Choi, Kiseok and Cho, Hyeongjun and Kim, Inchul and Kim, Min H.},
  booktitle = {Proceedings of the IEEE/CVF Conference on Computer Vision and Pattern Recognition},
  pages     = {25969--25978},
  month     = {June},
  year      = {2026}
}

@article{feldkamp1984practical,
  title={Practical cone-beam algorithm},
  author={Feldkamp, Lee A and Davis, Lloyd C and Kress, James W},
  journal={Journal of the Optical Society of America A},
  volume={1},
  number={6},
  pages={612--619},
  year={1984},
  publisher={Optical Society of America}
}

@article{gilbert1972iterative,
  title={Iterative methods for the three-dimensional reconstruction of an object from projections},
  author={Gilbert, Peter},
  journal={Journal of Theoretical Biology},
  volume={36},
  number={1},
  pages={105--117},
  year={1972},
  publisher={Elsevier}
}

@article{jin2017deep,
  title={Deep convolutional neural network for inverse problems in imaging},
  author={Jin, Kyong Hwan and McCann, Michael T and Froustey, Emmanuel and Unser, Michael},
  journal={IEEE Transactions on Image Processing},
  volume={26},
  number={9},
  pages={4509--4522},
  year={2017},
  publisher={IEEE}
}

@article{chen2018learn,
  title={LEARN: Learned experts’ assessment-based reconstruction network for sparse-data {CT}},
  author={Chen, Hu and Zhang, Yi and Chen, Yunjin and Zhang, Junfeng and Zhang, Weihua and Sun, Huaiqiang and Lv, Yang and Liao, Peixi and Zhou, Jiliu and Wang, Ge},
  journal={IEEE Transactions on Medical Imaging},
  volume={37},
  number={6},
  pages={1333--1347},
  year={2018},
  publisher={IEEE}
}

@inproceedings{chung2023solving,
  title={Solving {3D} inverse problems using pre-trained {2D} diffusion models},
  author={Chung, Hyungjin and Ryu, Dohoon and McCann, Michael T and Klasky, Marc L and Ye, Jong Chul},
  booktitle={Proceedings of the IEEE/CVF Conference on Computer Vision and Pattern Recognition},
  pages={22542--22551},
  year={2023},
  organization={IEEE}
}

@inproceedings{zha2022naf,
  title={{NAF}: neural attenuation fields for sparse-view {CBCT} reconstruction},
  author={Zha, Ruyi and Zhang, Yanhao and Li, Hongdong},
  booktitle={International Conference on Medical Image Computing and Computer-assisted Intervention},
  pages={442--452},
  year={2022},
  organization={Springer}
}

@inproceedings{cai2024structure,
  title={Structure-aware sparse-view {X}-ray {3D} reconstruction},
  author={Cai, Yuanhao and Wang, Jiahao and Yuille, Alan and Zhou, Zongwei and Wang, Angtian},
  booktitle={Proceedings of the IEEE/CVF Conference on Computer Vision and Pattern Recognition},
  pages={11174--11183},
  year={2024},
  organization={IEEE}
}

@article{thies2024gradient,
  title={A gradient-based approach to fast and accurate head motion compensation in cone-beam {CT}},
  author={Thies, Mareike and Wagner, Fabian and Maul, Noah and Yu, Haijun and Goldmann, Manuela and Schneider, Linda-Sophie and Gu, Mingxuan and Mei, Siyuan and Folle, Lukas and Preuhs, Alexander and others},
  journal={IEEE Transactions on Medical Imaging},
  volume={44},
  number={2},
  pages={1098--1109},
  year={2025},
  publisher={IEEE}
}

@inproceedings{wu2023joint,
  title={Joint rigid motion correction and sparse-view {CT} via self-calibrating neural field},
  author={Wu, Qing and Li, Xin and Wei, Hongjiang and Yu, Jingyi and Zhang, Yuyao},
  booktitle={2023 IEEE 20th International Symposium on Biomedical Imaging (ISBI)},
  pages={1--5},
  year={2023},
  organization={IEEE}
}

@article{der2019cone,
  title={A cone-beam X-ray computed tomography data collection designed for machine learning},
  author={Der Sarkissian, Henri and Lucka, Felix and Van Eijnatten, Maureen and Colacicco, Giulia and Coban, Sophia Bethany and Batenburg, Kees Joost},
  journal={Scientific Data},
  volume={6},
  number={1},
  pages={215},
  year={2019},
  publisher={Nature Publishing Group UK London}
}

@article{aichert2015epipolar,
  title={Epipolar consistency in transmission imaging},
  author={Aichert, Andr{\'e} and Berger, Martin and Wang, Jian and Maass, Nicole and Doerfler, Arnd and Hornegger, Joachim and Maier, Andreas K},
  journal={IEEE Transactions on Medical Imaging},
  volume={34},
  number={11},
  pages={2205--2219},
  year={2015},
  publisher={IEEE}
}

@article{gordon1970art,
  title={Algebraic reconstruction techniques (ART) for three-dimensional electron microscopy and X-ray photography},
  author={Gordon, Richard and Bender, Robert and Herman, Gabor T},
  journal={Journal of Theoretical Biology},
  volume={29},
  number={3},
  pages={471--481},
  year={1970},
  publisher={Elsevier}
}

@article{andersen1984sart,
  title={Simultaneous algebraic reconstruction technique ({SART}): a superior implementation of the {ART} algorithm},
  author={Andersen, Anders H and Kak, Avinash C},
  journal={Ultrasonic imaging},
  volume={6},
  number={1},
  pages={81--94},
  year={1984},
  publisher={Elsevier}
}

@article{beck2009fista,
  title={A fast iterative shrinkage-thresholding algorithm for linear inverse problems},
  author={Beck, Amir and Teboulle, Marc},
  journal={SIAM Journal on Imaging Sciences},
  volume={2},
  number={1},
  pages={183--202},
  year={2009},
  publisher={SIAM}
}

@article{adler2018learned,
  title={Learned primal-dual reconstruction},
  author={Adler, Jonas and {\"O}ktem, Ozan},
  journal={IEEE Transactions on Medical Imaging},
  volume={37},
  number={6},
  pages={1322--1332},
  year={2018},
  publisher={IEEE}
}

@article{shen2022nerp,
  title={{NeRP}: implicit neural representation learning with prior embedding for sparsely sampled image reconstruction},
  author={Shen, Liyue and Pauly, John and Xing, Lei},
  journal={IEEE Transactions on Neural Networks and Learning Systems},
  volume={35},
  number={1},
  pages={770--782},
  year={2024},
  publisher={IEEE}
}

@article{mildenhall2020nerf,
author = {Mildenhall, Ben and Srinivasan, Pratul P. and Tancik, Matthew and Barron, Jonathan T. and Ramamoorthi, Ravi and Ng, Ren},
title = {{NeRF}: representing scenes as neural radiance fields for view synthesis},
year = {2021},
publisher = {Association for Computing Machinery},
address = {New York, NY, USA},
volume = {65},
number = {1},
journal = {Communications of the ACM},
pages = {99–106}
}

@article{kerbl2023gaussian,
  title={{3D} {G}aussian splatting for real-time radiance field rendering.},
  author={Kerbl, Bernhard and Kopanas, Georgios and Leimk{\"u}hler, Thomas and Drettakis, George},
  journal={ACM Transactions on Graphics},
  volume={42},
  number={4},
  pages={139:1--139:14},
  year={2023}
}

@article{li2025threegrct,
  title={{3DGR-CT}: Sparse-view {CT} reconstruction with a {3D} {G}aussian representation},
  author={Li, Yingtai and Fu, Xueming and Li, Han and Zhao, Shang and Jin, Ruiyang and Zhou, S Kevin},
  journal={Medical Image Analysis},
  volume={103},
  pages={103585},
  year={2025},
  publisher={Elsevier}
}

@inproceedings{wu2025dgr,
  title={Discretized {G}aussian representation for tomographic reconstruction},
  author={Wu, Shaokai and Lu, Yuxiang and Guo, Yapan and Ji, Wei and Huang, Suizhi and Yang, Fengyu and Sirejiding, Shalayiding and He, Qichen and Tong, Jing and Ji, Yanbiao and others},
  booktitle={2025 IEEE/CVF International Conference on Computer Vision (ICCV)},
  pages={25073--25082},
  year={2025},
  organization={IEEE}
}

@inproceedings{pieta2026factgs,
  title={{Fact-GS}: Fast and scalable {CT} reconstruction with {G}aussian splatting},
  author={Pieta, Pawel Tomasz and Pedersen, Rasmus Juul and Borgi, Sina and J{\o}rgensen, Jakob Sauer and Andreasen, Jens Wenzel and Dahl, Vedrana Andersen},
  booktitle={European Conference on Computer Vision},
  pages={225--245},
  year={2026},
  organization={Springer Nature Switzerland}
}

@article{noo2000ellipse,
  title={Analytic method based on identification of ellipse parameters for scanner calibration in cone-beam tomography},
  author={Noo, Fr{\'e}d{\'e}ric and Clackdoyle, Rolf and Mennessier, Catherine and White, Timothy A and Roney, Timothy J},
  journal={Physics in Medicine \& Biology},
  volume={45},
  number={11},
  pages={3489--3508},
  year={2000}
}

@article{mennessier2009alignment,
  title={Direct determination of geometric alignment parameters for cone-beam scanners},
  author={Mennessier, Catherine and Clackdoyle, Rolf and Noo, Frederic},
  journal={Physics in Medicine \& Biology},
  volume={54},
  number={6},
  pages={1633--1660},
  year={2009}
}

@article{jacobson2018line,
  title={A line fiducial method for geometric calibration of cone-beam {CT} systems with diverse scan trajectories},
  author={Jacobson, MW and Ketcha, MD and Capostagno, S and Martin, A and Uneri, A and Goerres, J and De Silva, T and Reaungamornrat, S and Han, R and Manbachi, A and others},
  journal={Physics in Medicine \& Biology},
  volume={63},
  number={2},
  pages={025030},
  year={2018},
  publisher={IOP Publishing}
}

@article{luo2020misalignment,
  title={Data sustained misalignment correction in microscopic cone beam {CT} via optimization under the Grangeat Epipolar consistency condition},
  author={Luo, Shouhua and Zheng, Liang and Luo, Shuang and Gu, Ning and Tang, Xiangyang},
  journal={Medical Physics},
  volume={47},
  number={2},
  pages={498--508},
  year={2020},
  publisher={Wiley Online Library}
}

@article{wang2009mean,
  title={Mean squared error: Love it or leave it? A new look at signal fidelity measures},
  author={Wang, Zhou and Bovik, Alan C},
  journal={IEEE Signal Processing Magazine},
  volume={26},
  number={1},
  pages={98--117},
  year={2009},
  publisher={IEEE}
}

@article{wang2004image,
  title={Image quality assessment: from error visibility to structural similarity},
  author={Wang, Zhou and Bovik, Alan C and Sheikh, Hamid R and Simoncelli, Eero P},
  journal={IEEE Transactions on Image Processing},
  volume={13},
  number={4},
  pages={600--612},
  year={2004},
  publisher={IEEE}
}

@article{biguri2016tigre,
  title={{TIGRE}: a {MATLAB-GPU} toolbox for {CBCT} image reconstruction},
  author={Biguri, Ander and Dosanjh, Manjit and Hancock, Steven and Soleimani, Manuchehr},
  journal={Biomedical Physics \& Engineering Express},
  volume={2},
  number={5},
  pages={055010},
  year={2016},
  publisher={IOP Publishing}
}

@article{avants2011reproducible,
  title={A reproducible evaluation of {ANT}s similarity metric performance in brain image registration},
  author={Avants, Brian B and Tustison, Nicholas J and Song, Gang and Cook, Philip A and Klein, Arno and Gee, James C},
  journal={NeuroImage},
  volume={54},
  number={3},
  pages={2033--2044},
  year={2011},
  publisher={Elsevier}
}

@article{wasserthal2023totalsegmentator,
  title={{TotalSegmentator}: robust segmentation of 104 anatomic structures in {CT} images},
  author={Wasserthal, Jakob and Breit, Hanns-Christian and Meyer, Manfred T and Pradella, Maurice and Hinck, Daniel and Sauter, Alexander W and Heye, Tobias and Boll, Daniel T and Cyriac, Joshy and Yang, Shan and others},
  journal={Radiology: Artificial Intelligence},
  volume={5},
  number={5},
  pages={e230024},
  year={2023},
  publisher={Radiological Society of North America}
}

@inproceedings{lin2021barf,
  title={{Barf}: Bundle-adjusting neural radiance fields},
  author={Lin, Chen-Hsuan and Ma, Wei-Chiu and Torralba, Antonio and Lucey, Simon},
  booktitle={2021 IEEE/CVF International Conference on Computer Vision},
  pages={5721--5731},
  year={2021},
  organization={IEEE}
}

@inproceedings{fu2024colmap,
  title={Colmap-free {3D} {G}aussian splatting},
  author={Fu, Yang and Liu, Sifei and Kulkarni, Amey and Kautz, Jan and Efros, Alexei A and Wang, Xiaolong},
  booktitle={Proceedings of the IEEE/CVF Conference on Computer Vision and Pattern Recognition},
  pages={20796--20805},
  year={2024}
}

@inproceedings{jeong2021self,
  title={Self-calibrating neural radiance fields},
  author={Jeong, Yoonwoo and Ahn, Seokjun and Choy, Christopher and Anandkumar, Anima and Cho, Minsu and Park, Jaesik},
  booktitle={Proceedings of the IEEE/CVF International Conference on Computer Vision},
  pages={5846--5854},
  year={2021}
}

@inproceedings{grangeat1991framework,
  title={Mathematical framework of cone beam {3D} reconstruction via the first derivative of the {R}adon transform},
  author={Grangeat, Pierre},
  booktitle={Mathematical Methods in Tomography: Proceedings of a Conference held in Oberwolfach, Germany, 5--11 June, 1990},
  pages={66--97},
  year={2006},
  organization={Springer}
}
}

\clearpage
\setcounter{page}{1}
\maketitlesupplementary

\section{Differentiable Epipolar Consistency Formulation}
\label{supp_ecc}
The epipolar consistency conditions between two cone-beam projections follow from Grangeat's theorem~\cite{grangeat1991framework} and were formulated for transmission imaging by Aichert \etal~\cite{aichert2015epipolar}. Here we describe how we evaluate the term $\mathcal{C}_k(E;\tilde{\gamma})$ of Eq.~\eqref{eq:ecc_loss} as a differentiable function of the scanner parameters $\tilde{\gamma}$.

\noindent\textbf{Acquisition geometry.}
From $\tilde{\gamma}=(\delta u,\delta v,\eta,\phi,\theta)$ (Eq.~\eqref{eq7}) we recover the physical offsets $(u,v)$ by
Eq.~\eqref{eq:residual_geometry}, giving $\gamma=(u,v,\eta,\phi,\theta)$. For view $k$ with view angle $\beta_k$, the source is $\mathbf{s}_k=R\,(\cos\beta_k,\sin\beta_k,0)^\top$, with $R$ the source-to-isocenter distance. Since the detector tilts about the source, $\gamma$ determines only the detector centre $\mathbf{o}_k(\gamma)$ and the orthonormal detector basis $\{\mathbf{e}_{k,u},\mathbf{e}_{k,v},\mathbf{e}_{k,n}\}(\gamma)$, where $\mathbf{e}_{k,u}$ and $\mathbf{e}_{k,v}$ are the horizontal and vertical detector axes and $\mathbf{e}_{k,n}=\mathbf{e}_{k,u}\times \mathbf{e}_{k,v}$ is the detector normal. A detector point with metric coordinates $(\xi,\zeta)$ is $\mathbf{P}=\mathbf{o}_k+\xi\,\mathbf{e}_{k,u}+\zeta\,\mathbf{e}_{k,v}$, and $y_k(\mathbf{P})$ denotes the measured projection of view $k$ at $\mathbf{P}$.

\noindent\textbf{Epipolar planes.}
For a pair $(i,j)\in\Omega$, the planes containing both sources form a pencil around the
baseline~\cite{aichert2015epipolar}. Let $\hat{\mathbf{b}}=(\mathbf{s}_j-\mathbf{s}_i)/\|\mathbf{s}_j-
\mathbf{s}_i\|$ be the baseline direction, $\mathbf{a}_1$ the unit vector obtained by orthogonalizing the rotation
axis $\mathbf{e}_z$ against $\hat{\mathbf{b}}$, and $\mathbf{a}_2=\hat{\mathbf{b}}\times\mathbf{a}_1$. The planes are
then $E(\kappa)=\{\mathbf{x}:\mathbf{n}_E\cdot\mathbf{x}=\rho_E\}$ with
\begin{equation}
  \mathbf{n}_E(\kappa)=\cos\kappa\,\mathbf{a}_1+\sin\kappa\,\mathbf{a}_2,
  \rho_E=\mathbf{n}_E\cdot\mathbf{s}_i=\mathbf{n}_E\cdot\mathbf{s}_j,
  \label{eq:supp_ecc_plane}
\end{equation}
where $\kappa$ is the rotation angle of the plane about the baseline ($\kappa=0$: the orbit plane) and $\rho_E$
is the signed distance of $E$ from the isocenter. Since the sources do not depend on $\tilde{\gamma}$, these planes are fixed in world coordinates; $\tilde{\gamma}$ only determines where each plane meets the two detectors. As $\mathbf{s}_k\in E$, a detector point $\mathbf{P}$ of view $k\in\{i,j\}$ lies in $E$ if and only if $(\mathbf{P}-\mathbf{s}_k)\cdot \mathbf{n}_E=0$, which gives the epipolar line
\begin{equation}
  \underbrace{(\mathbf{e}_{k,u}\!\cdot\!\mathbf{n}_E)}_{a_k}\,\xi
  +\underbrace{(\mathbf{e}_{k,v}\!\cdot\!\mathbf{n}_E)}_{b_k}\,\zeta
  =\underbrace{(\mathbf{s}_k-\mathbf{o}_k)\!\cdot\!\mathbf{n}_E}_{c_k}.
  \label{eq:supp_ecc_line}
\end{equation}
We parameterize it by arc length $\tau$,
\begin{equation}
  \mathbf{P}_k(\tau)=\mathbf{o}_k+\frac{c_k\,(a_k\mathbf{e}_{k,u}+b_k\mathbf{e}_{k,v})}{a_k^2+b_k^2}
  +\tau\,\mathbf{t}_k,
  \label{eq:supp_ecc_param}
\end{equation}
where the unit vector $\mathbf{t}_k\propto-b_k\mathbf{e}_{k,u}+a_k\mathbf{e}_{k,v}$ is the direction of the line.

\noindent\textbf{Grangeat's relation.}
Let $f$ be the attenuation field and
\begin{equation}
  X_k(\boldsymbol{\omega})=\int_0^\infty f(\mathbf{s}_k+t\,\boldsymbol{\omega})\,dt
  =y_k\big(\mathbf{P}_k(\boldsymbol{\omega})\big)
  \label{eq:supp_ecc_X}
\end{equation}
the line integral of view $k$ along the ray from $\mathbf{s}_k$ in unit direction $\boldsymbol{\omega}$, where
$\mathbf{P}_k(\boldsymbol{\omega})$ is the point at which this ray hits the detector. Tilting every ray of view $k$
in $E$ towards the plane normal by a small angle $\varepsilon$, $\boldsymbol{\omega}_\varepsilon=\boldsymbol{\omega}+
\varepsilon\,\mathbf{n}_E$, Grangeat's theorem gives~\cite{grangeat1991framework,aichert2015epipolar}
\begin{equation}
  \int\frac{\partial X_k}{\partial\varepsilon}\,d\psi
  =\int_E\nabla f\cdot\mathbf{n}_E\,dA
  =\frac{\partial}{\partial\rho}\mathcal{R}f(\mathbf{n}_E,\rho)\Big|_{\rho=\rho_E},
  \label{eq:supp_ecc_grangeat}
\end{equation}
where $\psi$ is the fan angle in $E$ and $\mathcal{R}f(\mathbf{n},\rho)$ is the integral of $f$ over the plane
$\{\mathbf{x}:\mathbf{n}\cdot\mathbf{x}=\rho\}$. In polar coordinates about $\mathbf{s}_k$ in $E$, with $t$ the distance from the source, $dA=t\,dt\,d\psi$. The plain fan sum $\int X_k\,d\psi=\int_E f/t\,dA$ therefore weights each point of $E$ by its inverse
distance to the source, which differs between the two views~\cite{aichert2015epipolar}.
Tilting a ray by $\varepsilon$ moves its point at distance $t$ by $t\varepsilon$, so
$\partial X_k/\partial\varepsilon=\int t\,\nabla f(\mathbf{s}_k+t\boldsymbol{\omega})\cdot\mathbf{n}_E\,dt$;
this factor $t$ cancels the weight $1/t$ and yields the unweighted integral in Eq.~\eqref{eq:supp_ecc_grangeat}. The right-hand side therefore depends only
on $E$, and the two views of a pair must give the same value, provided that the fan covers the whole intersection of
the object with $E$ (no truncation along the detector line).

\noindent\textbf{Evaluation on the detector.}
We evaluate the left-hand side of Eq.~\eqref{eq:supp_ecc_grangeat} on the measured projection $y_k$, with the
detector frame and the epipolar line computed from $\tilde{\gamma}$. The tilted ray
$\mathbf{s}_k+t\,\boldsymbol{\omega}_\varepsilon$ meets the detector plane $\{\mathbf{x}:(\mathbf{x}-\mathbf{o}_k)
\cdot\mathbf{e}_{k,n}=0\}$ at $t(\varepsilon)=(\mathbf{o}_k-\mathbf{s}_k)\cdot\mathbf{e}_{k,n}/
(\boldsymbol{\omega}_\varepsilon\cdot\mathbf{e}_{k,n})$. With $\mathbf{r}=\mathbf{P}-\mathbf{s}_k$,
$\ell=\|\mathbf{r}\|$ and $\boldsymbol{\omega}=\mathbf{r}/\ell$, we have $t(0)=\ell$ and
$t'(0)=-\ell\,(\mathbf{n}_E\cdot\mathbf{e}_{k,n})/(\boldsymbol{\omega}\cdot\mathbf{e}_{k,n})$, so the hit point
$\mathbf{P}(\varepsilon)=\mathbf{s}_k+t(\varepsilon)\,\boldsymbol{\omega}_\varepsilon$ moves with velocity
\begin{equation}
  \frac{d\mathbf{P}}{d\varepsilon}=t(0)\,\mathbf{n}_E+t'(0)\,\boldsymbol{\omega}
  =\ell\Big(\mathbf{n}_E-\frac{\mathbf{n}_E\cdot\mathbf{e}_{k,n}}{\boldsymbol{\omega}\cdot\mathbf{e}_{k,n}}\,
  \boldsymbol{\omega}\Big),
  \label{eq:supp_ecc_dP}
\end{equation}
which lies in the detector plane. The chain rule applied to $X_k=y_k\circ\mathbf{P}_k$ in Eq.~\eqref{eq:supp_ecc_X}
gives
\begin{equation}
  \frac{\partial X_k}{\partial\varepsilon}=\nabla_d\,y_k\cdot\mathbf{g}_k,
  \mathbf{g}_k=\Big(\frac{d\mathbf{P}}{d\varepsilon}\cdot\mathbf{e}_{k,u},\;
  \frac{d\mathbf{P}}{d\varepsilon}\cdot\mathbf{e}_{k,v}\Big),
  \label{eq:supp_ecc_chain}
\end{equation}
where $\nabla_d y_k=(\partial y_k/\partial\xi,\,\partial y_k/\partial\zeta)$ is the gradient of the projection in
the detector plane. Along the line $\mathbf{P}_k(\tau)$, the ray direction turns at the rate
$\|d\boldsymbol{\omega}/d\tau\|=\|\mathbf{r}\times\mathbf{t}_k\|/\ell^2$, so $d\psi=\|\mathbf{r}\times
\mathbf{t}_k\|/\ell^2\,d\tau$. Altogether, the Grangeat-domain quantity of view $k$ under the geometry $\tilde{\gamma}$ is
\begin{equation}
  \mathcal{C}_k(E;\tilde{\gamma})=\int\nabla_d\,y_k\big(\mathbf{P}_k(\tau)\big)\cdot\mathbf{g}_k(\tau)\,
  \frac{\|\mathbf{r}(\tau)\times\mathbf{t}_k\|}{\ell(\tau)^2}\,d\tau ,
  \label{eq:supp_ecc_final}
\end{equation}
a weighted line integral of the projection gradient along the epipolar line, in which $\tilde{\gamma}$ enters
through the line position $\mathbf{P}_k(\tau)$, the detector basis, $\mathbf{g}_k$ and the weight (all computed from
$\tilde{\gamma}$ by Eqs.~\eqref{eq:supp_ecc_line}--\eqref{eq:supp_ecc_chain}). Setting $k=i$ and $k=j$ gives the two
terms $\mathcal{C}_i(E;\tilde{\gamma})$ and $\mathcal{C}_j(E;\tilde{\gamma})$ of Eq.~\eqref{eq:ecc_loss}. If $\tilde{\gamma}$ is the true geometry, both equal $\partial_\rho\mathcal{R}f(\mathbf{n}_E,\rho_E)$ by
Eq.~\eqref{eq:supp_ecc_grangeat}, so $\mathcal{C}_i(E;\tilde{\gamma})=\mathcal{C}_j(E;\tilde{\gamma})$ for every
plane through both sources, up to discretization. If $\tilde{\gamma}$ is wrong, the lines of
Eq.~\eqref{eq:supp_ecc_line} computed in the two views no longer belong to the same physical plane, and the two values generally
differ; $\mathcal{L}_{\text{ECC}}$ penalizes this difference.

\noindent\textbf{Implementation.}
The measured projections are smoothed once by a Gaussian ($\sigma=2$\,px). The candidate pairs are all view pairs
at least $5^\circ$ apart (at most 1500). For each pair, 64 planes uniformly spanning the range of $\kappa$ whose
detector lines stay within $90\,\%$ of the detector height across the full detector width on both views are
selected once at the initial geometry and kept fixed, so that the plane integrals are not truncated and the
optimizer cannot lower the cost by moving the planes; they form the plane set $\mathcal{E}_{ij}$ of Eq.~\eqref{eq:ecc_loss}. In
each iteration, $|\Omega|=128$ pairs are drawn at random; along each line, the projection and its central-difference
gradient (1\,px) are sampled every 2\,px by differentiable bilinear interpolation, and Eq.~\eqref{eq:supp_ecc_final} is evaluated as a Riemann sum, so that gradients reach $\tilde{\gamma}$ through the
line positions, $\mathbf{g}_k$ and $d\psi$. The loss is the mean of $(\mathcal{C}_i-\mathcal{C}_j)^2$ over the
sampled pairs and planes, divided by a constant fixed at the initial geometry (the mean of
$(\mathcal{C}_i^2+\mathcal{C}_j^2)/2$ over all pairs), which makes $\lambda_{\mathrm{ECC}}=1$ independent of the
intensity scale. Following the identifiability analysis (Fig.~\ref{fig1}(d)), $\theta$ enters $\mathcal{L}_{\text{ECC}}$ as a
constant (stop-gradient), so that $\mathcal{L}_{\text{ECC}}$ constrains $(\delta u,\delta v,\eta,\phi)$ only.
$\mathcal{L}_{\text{ECC}}$ is used during the first 5000 joint iterations; afterwards $\delta v$ is frozen and the
other parameters are refined by projection matching alone.

\section{Additional Experimental Details}

\subsection{Simulated Datasets}
We used the head and abdomen CT volumes of size $256^3$ from the R$^2$-Gaussian benchmark~\cite{r2_gaussian}. Cone-beam projections were simulated on a $512^2$ detector. Poisson noise with $I_0=10^5$ and additive Gaussian noise with a standard deviation of 10 photon counts were included in the simulated measurements.
For the standard sparse-view setting, 50 views uniformly distributed over $360^\circ$ were used for reconstruction, and another 100 views were reserved exclusively for evaluation. All calibration methods started from the same misaligned nominal geometry.

\subsection{Real-world Datasets and Acquisition Systems}

Our main real-data experiments used the CWI FleX-ray walnut dataset~\cite{der2019cone} and corn-kernel acquired using our laboratory-built CBCT system. The laboratory system acquired 1024 projections with SOD/SDD of 200/335 mm.
We additionally evaluated whether the scanner-level geometry estimated from a sparse subset could be reused for conventional full-view reconstruction. An optical-fibre bundle and a carbon-fibre composite were scanned using an industrial micro-CT system with SOD/SDD of 22/600 mm and 40/600 mm, respectively. A bean sample was scanned using the laboratory CBCT system. In each case, the geometry was estimated from 50 projections and subsequently applied to FDK reconstruction using all available views.

\subsection{Evaluation}
\noindent\textbf{3D Reconstruction Accuracy.}
Before computing 3D image-quality metrics, each reconstructed volume was rigidly registered to the corresponding reference using a 6-DoF transformation~\cite{avants2011reproducible}. This registration avoids penalizing global rigid offsets that are intrinsically ambiguous with some acquisition-geometry parameters. In particular, a vertical detector displacement can be partially compensated by translation of the reconstructed object along the rotation axis.

\noindent\textbf{Held-Out Projection Evaluation.}
For real data, we additionally evaluated each reconstruction using 100 measured projections that were completely excluded from calibration and reconstruction. The reconstructed volume was projected onto these views using the estimated geometry, and PSNR/SSIM were computed between the predicted and measured projections, following the held-out-view evaluation strategy used in self-calibrating reconstruction~\cite{ruckert2022neat}. This evaluation measures predictive consistency on unseen views, although high agreement alone does not establish the accuracy of the recovered physical geometry because geometry–volume ambiguity may persist. Panetta \etal~\cite{panetta2008optimization} and Haouchat \etal~\cite{haouchat2026geometry} estimate a shared scanner geometry. For these methods, R$^2$-Gaussian was retrained using the estimated geometry before held-out evaluation, so that the Gaussian reconstruction pipeline was kept consistent. NeAT~\cite{ruckert2022neat} and Choi \etal~\cite{choi2026pose} instead estimate independent per-view poses. Since no correction is directly available for unseen views, the nominal held-out poses were corrected using the mean pose correction estimated from the training views.

\begin{table}[t]
\centering
\caption{Ours (first row of each condition, as Tab.~\ref{tab:extreme}) and the ECC-only calibration (second row)
under extreme conditions, both followed by the same fixed-geometry reconstruction as ours (abdomen, $\times4$ error). Absolute geometry errors
($|\Delta u|,|\Delta v|$ in px; $|\eta|,|\phi|,|\theta|$ in degrees) and registered 3D PSNR (dB)/SSIM of the final
reconstruction. Standard: 50 views over $360^\circ$, $I_0{=}10^5$. $^\dagger$At the $\pm5^\circ$ search bound.}
\label{tab:supp_ecconly_extreme}
\setlength{\tabcolsep}{2.6pt}
\setlength{\aboverulesep}{0pt}
\setlength{\belowrulesep}{0pt}
\renewcommand{\arraystretch}{1.12}
\footnotesize
\begin{tabular}{l ccccc cc}
\toprule
Condition & $|\Delta u|$ & $|\Delta v|$ & $|\eta|$ & $|\phi|$ & $|\theta|$ & PSNR & SSIM \\
\midrule
Standard                    & 0.012 & 0.16 & 0.001 & 0.000 & 0.69 & 35.80 & 0.951 \\
\quad ECC-only              & 0.005 & 0.34 & 0.001 & 0.005 & 3.84 & 31.07 & 0.923 \\
\midrule
$I_0{=}10^4$                & 0.017 & 0.43 & 0.002 & 0.001 & 0.62 & 35.16 & 0.936 \\
\quad ECC-only              & 0.001 & 1.03 & 0.005 & 0.014 & 5.00$^\dagger$ & 29.72 & 0.900 \\
$I_0{=}2{\times}10^3$       & 0.021 & 0.90 & 0.000 & 0.004 & 1.09 & 32.18 & 0.872 \\
\quad ECC-only              & 0.058 & 3.22 & 0.009 & 0.082 & 5.00$^\dagger$ & 28.31 & 0.845 \\
\midrule
20 views                    & 0.002 & 0.14 & 0.001 & 0.006 & 0.38 & 30.04 & 0.876 \\
\quad ECC-only              & 0.028 & 0.07 & 0.004 & 0.013 & 3.27 & 28.69 & 0.860 \\
10 views                    & 0.008 & 0.05 & 0.000 & 0.043 & 0.75 & 27.07 & 0.811 \\
\quad ECC-only              & 0.063 & 0.09 & 0.000 & 0.006 & 0.96 & 26.95 & 0.809 \\
\bottomrule
\end{tabular}
\end{table}

\begin{table}[t]
\centering
\caption{Absolute errors of the estimated scanner geometry at the intermediate error level $\times2$ (Tab.~\ref{tab:supp_levels}), mean over head and abdomen ($|\Delta u|,|\Delta v|$ in px; $|\eta|,|\phi|,|\theta|$ in degrees; residual coordinates as in Tab.~\ref{tab:main}). $^\dagger$Not estimated (nominal value kept). $^\ddagger$Per-view poses: the global geometry that best reproduces the mean learned camera correction. \colorbox{bestbg}{\textbf{Best}} and \colorbox{secondbg}{second best} results are highlighted.}
\label{tab:supp_levels_geo}
\setlength{\tabcolsep}{4pt}
\setlength{\aboverulesep}{0pt}
\setlength{\belowrulesep}{0pt}
\renewcommand{\arraystretch}{1.15}
\small
\begin{tabular}{l ccccc}
\toprule
Method & $|\Delta u|$ & $|\Delta v|$ & $|\eta|$ & $|\phi|$ & $|\theta|$ \\
\midrule
\textcolor{gray}{Initial error} & \textcolor{gray}{19.640} & \textcolor{gray}{11.64} & \textcolor{gray}{1.000} & \textcolor{gray}{1.000} & \textcolor{gray}{1.00} \\
\midrule
Panetta \etal~\cite{panetta2008optimization} & 0.115 & 11.64$^\dagger$ & 0.091 & 0.388 & 1.00$^\dagger$ \\
Haouchat \etal~\cite{haouchat2026geometry} & \csecond{0.072} & \csecond{5.74} & \csecond{0.004} & 0.221 & \csecond{0.68} \\
NeAT$^\ddagger$~\cite{ruckert2022neat} & 1.513 & 10.41 & 1.014 & 0.856 & 1.06 \\
Choi \etal$^\ddagger$~\cite{choi2026pose} & 0.135 & 7.63 & 0.018 & \csecond{0.090} & 0.77 \\
Ours & \cbest{0.005} & \cbest{0.15} & \cbest{0.000} & \cbest{0.004} & \cbest{0.47} \\
\bottomrule
\end{tabular}
\end{table}

\begin{table*}[t]
\centering
\caption{Intermediate error level $\times2$: registered 3D PSNR (dB)/SSIM on head and abdomen; the $\times1$ and $\times4$ columns are those of Tab.~\ref{tab:main}. Geometry errors at $\times2$: Tab.~\ref{tab:supp_levels_geo}. \colorbox{bestbg}{\textbf{Best}} and \colorbox{secondbg}{second best} results are highlighted.}
\label{tab:supp_levels}
\setlength{\tabcolsep}{4pt}
\setlength{\aboverulesep}{0pt}
\setlength{\belowrulesep}{0pt}
\renewcommand{\arraystretch}{1.15}
\small
\begin{tabular}{l cccccc}
\toprule
\multirow{2}{*}{Method} & \multicolumn{3}{c}{Head (PSNR/SSIM)} & \multicolumn{3}{c}{Abdomen (PSNR/SSIM)} \\
\cmidrule(lr){2-4}\cmidrule(lr){5-7}
& $\times1$ & $\times2$ & $\times4$ & $\times1$ & $\times2$ & $\times4$ \\
\midrule
FDK~\cite{feldkamp1984practical} & 23.69/0.375 & 21.31/0.345 & 19.14/0.314 & 21.07/0.259 & 19.47/0.235 & 17.84/0.186 \\
SIRT~\cite{gilbert1972iterative} & 24.70/0.648 & 21.13/0.542 & 18.80/0.396 & 22.14/0.591 & 19.82/0.506 & 17.86/0.386 \\
R$^2$-Gaussian~\cite{r2_gaussian} & 25.44/0.778 & 21.76/0.681 & 18.63/0.549 & 23.72/0.738 & 21.42/0.669 & 18.68/0.596 \\
\midrule
Panetta \etal~\cite{panetta2008optimization} & 28.34/0.488 & 27.20/0.482 & 25.57/0.486 & 25.21/0.376 & 24.49/0.374 & 23.23/0.373 \\
Haouchat \etal~\cite{haouchat2026geometry} & 31.42/0.707 & 28.72/0.674 & 27.05/0.658 & 28.19/0.571 & 27.48/0.564 & 26.60/0.546 \\
NeAT~\cite{ruckert2022neat} & 30.39/0.867 & 22.20/0.542 & 17.20/0.398 & 26.19/0.780 & 19.27/0.495 & 17.13/0.488 \\
Choi \etal~\cite{choi2026pose} & 36.71/0.966 & 33.47/0.956 & 30.25/0.935 & 32.13/0.930 & 29.99/0.918 & 27.67/0.892 \\
Ours (w/o refinement) & \csecond{39.00/0.973} & \csecond{38.66/0.971} & \csecond{37.95/0.967} & \csecond{36.27/0.951} & \csecond{35.65/0.947} & \csecond{34.81/0.941} \\
Ours & \cbest{39.70/0.976} & \cbest{39.64/0.976} & \cbest{39.49/0.975} & \cbest{36.65/0.953} & \cbest{36.07/0.951} & \cbest{35.80/0.951} \\
\bottomrule
\end{tabular}
\end{table*}

\noindent\textbf{Downstream Segmentation.}
To evaluate anatomical fidelity beyond pixel-wise image-quality metrics, we performed downstream segmentation using TotalSegmentator~\cite{wasserthal2023totalsegmentator}. Dice scores were averaged over 30 anatomical structures for the head volume and 40 structures for the abdomen volume.

\noindent\textbf{Geometry Accuracy and Runtime.}
For simulated experiments, geometry accuracy was measured using the absolute errors of the five estimated detector parameters $(u,v,\eta,\phi,\theta)$. Detector offsets are reported in pixels and detector rotations in degrees.
Run time was measured on a single NVIDIA RTX A6000 using the abdomen $\times1$ setting.

\subsection{Implementation Details}

CalibGS was implemented on top of R$^2$-Gaussian~\cite{r2_gaussian} and consisted  of two optimization stages.

\noindent\textbf{Joint Calibration.}
Joint calibration was optimized for 14k iterations using a fixed set of 50k Gaussian primitives. Gaussian densification was disabled during this stage to prevent an increasingly flexible Gaussian representation from absorbing geometric errors. 
The first 500 iterations were used as a warm-up stage. After warm-up, the Gaussian representation and the five geometry parameters were jointly optimized using Adam. The initial geometry learning rates were $0.02$ px for detector offsets and $0.002^\circ$ for angular parameters, followed by exponential decay. The detector pitch $\theta$ was frozen during the first 3000 iterations of geometry optimization. After being released, it was optimized only by the projection-matching term, with the ECC gradient detached. Because the projection sensitivity with respect to $\theta$ is weak, its angular update step was enlarged by a factor of 10. The epipolar-consistency loss followed Aichert \etal~\cite{aichert2015epipolar}. At each iteration, 128 projection pairs were randomly sampled from the candidate projection pairs. We used $\lambda_{\mathrm{ECC}}=1$ and $\lambda_{\mathrm{D\text{-}SSIM}}=0.05$. To perform analytical reconstruction using all five calibrated detector parameters, we extended the TIGRE projector~\cite{biguri2016tigre} to incorporate the corresponding detector offsets and rotations.

\noindent\textbf{Fixed-Geometry Reconstruction.}
After joint calibration, the calibrated geometry was fixed. The Gaussian representation was re-initialized from an FDK reconstruction generated using the calibrated geometry, and the standard R$^2$-Gaussian pipeline~\cite{r2_gaussian} was subsequently optimized for another 14k iterations with its normal reconstruction capacity and densification strategy.
This fixed-geometry reconstruction prevents errors introduced during early reconstruction under inaccurate geometry from remaining in the final Gaussian representation.

\subsection{Baseline Implementation Details}
We compared CalibGS with FDK~\cite{feldkamp1984practical}, SIRT~\cite{gilbert1972iterative}, R$^2$-Gaussian~\cite{r2_gaussian}, Panetta \etal~\cite{panetta2008optimization}, Haouchat \etal~\cite{haouchat2026geometry}, NeAT~\cite{ruckert2022neat}, and Choi \etal~\cite{choi2026pose}. All methods started from the same nominal geometry.

\noindent\textbf{FDK, SIRT, and R$^2$-Gaussian.}
FDK, SIRT, and R$^2$-Gaussian directly used the nominal geometry without calibration. SIRT was run for 300 iterations.

\noindent\textbf{Panetta \etal.}
No public implementation of Panetta \etal~\cite{panetta2008optimization} was available. We therefore re-implemented the method following the original formulation.

\noindent\textbf{Haouchat \etal.}
Haouchat \etal~\cite{haouchat2026geometry} jointly optimize the image representation and acquisition geometry. The 2D version of this method uses a B-spline representation. We therefore used its quadratic B-spline formulation and differentiated the projection operation through dense sampling along the rays.

\noindent\textbf{NeAT.}
NeAT~\cite{ruckert2022neat} was evaluated using the authors' released implementation. Its pose optimization was applied without changing the underlying per-view pose parameterization.

\noindent\textbf{Choi \etal.}
We evaluated the method of Choi \etal~\cite{choi2026pose} using the authors' released implementation. The original pose learning rate produced only small pose updates under the systematic errors considered in our experiments. We therefore tested several learning rates and reported the best-performing setting: $10^{-3}$ for the standard experiments and $4\times10^{-3}$ for the $\times4$ and Corn experiments.

\section{ECC-only Calibration}
\label{sec:supp_ecconly}
For the ECC-only setting of Tab.~\ref{tab:supp_ecconly_extreme}, no Gaussians were used: the five parameters were estimated by minimizing $\mathcal{L}_{\mathrm{ECC}}$ alone, with the same smoothing and epipolar planes as in Ours (w/o refinement) but all candidate pairs (all view pairs at least $5^\circ$ apart, at most 1500) in every evaluation, by bounded search over $\tilde{\gamma}$ ($\pm40$ and $\pm48$\,px around the initial $\delta u$ and $\delta v$; $|\eta|\le5^\circ$, $|\phi|\le8^\circ$, $|\theta|\le5^\circ$), restarted until the cost stopped decreasing. The resulting geometry was then fixed and used in the same fixed-geometry reconstruction as ours. Since $\mathcal{L}_{\mathrm{ECC}}$ is nearly flat along $\theta$, its minimum lay at $\theta=-3.8^\circ$ in the standard setting.
Table~\ref{tab:supp_ecconly_extreme} compares the ECC-only calibration with Ours under the extreme conditions of Tab.~\ref{tab:extreme}, both followed by the same fixed-geometry reconstruction as ours. ECC-only calibration yields larger pitch errors and lower reconstruction PSNR/SSIM than our method across all evaluated conditions.

\section{Additional simulated experiments}
\label{sec:supp_sim}
\begin{figure*}[t]
\centering
\includegraphics[width=\textwidth]{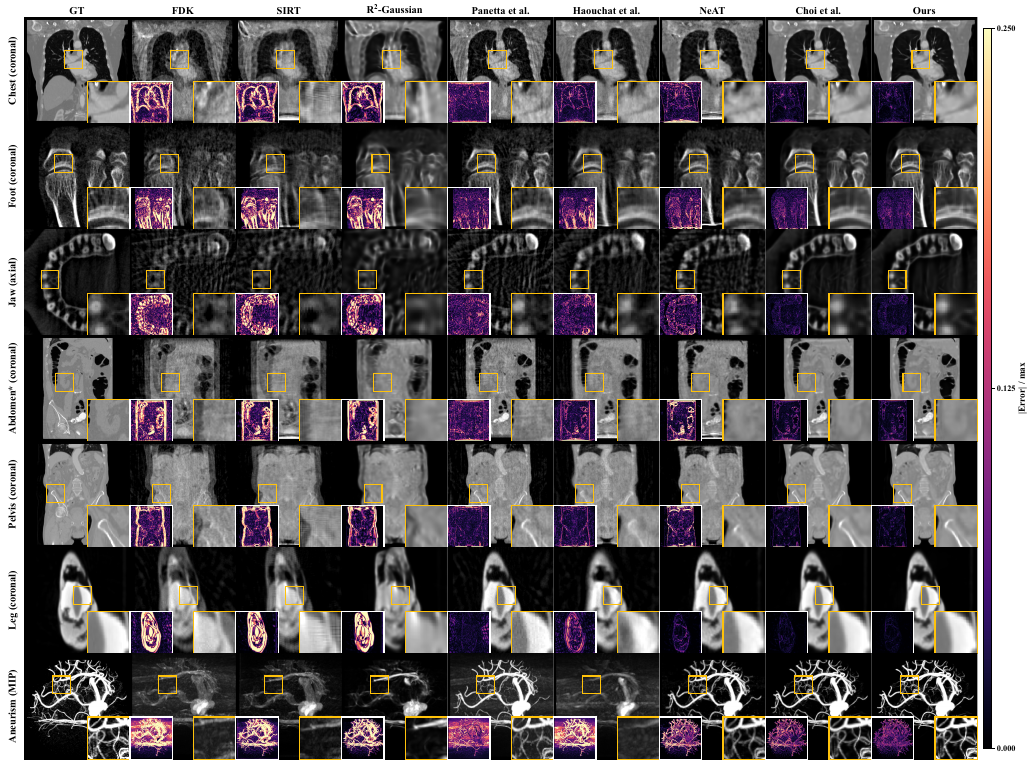}
\caption{Visual comparison on the seven further objects of Tab.~\ref{tab:supp_objects}, small error ($\times1$).
One view per object, the same for every method and both error levels: central coronal slice (chest, foot, abdomen,
pelvis, leg), central axial slice (jaw) and maximum intensity projection (MIP; aneurism, whose vessels appear only as
scattered dots in a single slice). Volumes rigidly registered to the ground truth (GT); display window $[0, p_{99.9}]$,
with $p_{99.9}$ the 99.9th percentile of the GT volume (of the GT MIP for the aneurism). Bottom right: zoom of the
yellow box; bottom left: $|\text{reconstruction}-\text{GT}|/p_{99.9}$ (these volumes have no HU calibration). $^*$A different patient from the abdomen of the main paper.}
\label{fig:supp_objects_x1}
\end{figure*}

\begin{figure*}[t]
\centering
\includegraphics[width=\textwidth]{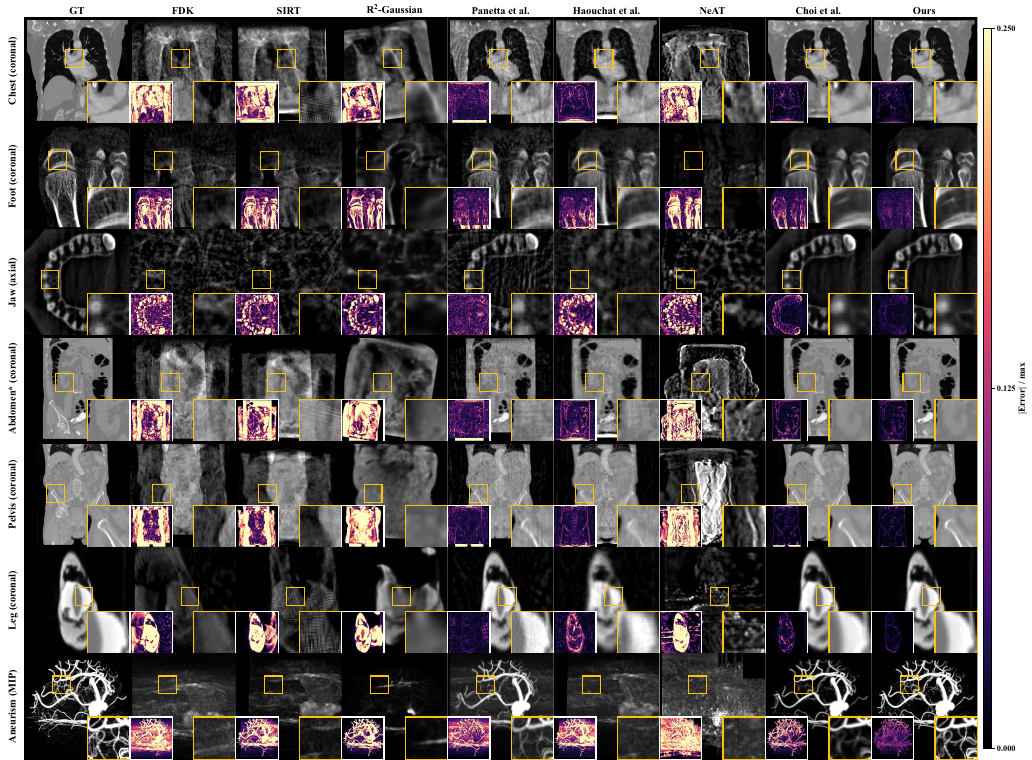}
\caption{Visual comparison on the seven further objects of Tab.~\ref{tab:supp_objects}, large error ($\times4$).
One view per object, the same for every method and both error levels: central coronal slice (chest, foot, abdomen,
pelvis, leg), central axial slice (jaw) and maximum intensity projection (MIP; aneurism, whose vessels appear only as
scattered dots in a single slice). Volumes rigidly registered to the ground truth (GT); display window $[0, p_{99.9}]$,
with $p_{99.9}$ the 99.9th percentile of the GT volume (of the GT MIP for the aneurism). Bottom right: zoom of the
yellow box; bottom left: $|\text{reconstruction}-\text{GT}|/p_{99.9}$ (these volumes have no HU calibration). $^*$A different patient from the abdomen of the main paper.}
\label{fig:supp_objects_x4}
\end{figure*}

The main experiments use two objects (head and abdomen) and a single direction of the initial geometry error at two magnitudes ($\times1$, $\times4$). Here we test whether the results carry over to other anatomies (Tab.~\ref{tab:supp_objects}, Figs.~\ref{fig:supp_objects_x1} and~\ref{fig:supp_objects_x4}), to an intermediate magnitude (Tab.~\ref{tab:supp_levels}) and to random error directions (Tab.~\ref{tab:supp_random}). Unless stated otherwise, the simulation, the error levels, the evaluation and the settings of all methods are those of Tab.~\ref{tab:main}. As in Tab.~\ref{tab:main}, geometry errors are given in the residual coordinates of Eq.~\eqref{eq:residual_geometry}: $|\Delta u|$ and $|\Delta v|$ are the errors of $\delta u$ and $\delta v$, not of the physical offsets $u$ and $v$.

\noindent\textbf{Additional CT Volumes.}
We add seven objects: the chest, foot, jaw, abdomen, pelvis, leg and aneurism CT volumes, which we resampled to $256^3$ voxels and re-simulated with our protocol (50 views over $360^\circ$, same scanner and noise model). This abdominal CT volume was acquired from a different patient than the one used in the main experiments. Our method gave the best reconstruction in all 14 cases (Tab.~\ref{tab:supp_objects}). The strongest baseline was Choi \etal in every case; averaged over the seven objects, ours reached 38.16\,dB against their 34.87\,dB at $\times1$, and 36.89\,dB against 30.14\,dB at $\times4$. From $\times1$ to $\times4$, our mean PSNR dropped by 1.3\,dB, that of Choi \etal by 4.7\,dB and that of NeAT by 9.5\,dB. At $\times1$, our reconstructions were within 0.25\,dB of R$^2$-Gaussian with the true geometry on average (38.16 vs.\ 38.41\,dB). Figs.~\ref{fig:supp_objects_x1} and~\ref{fig:supp_objects_x4} show one visualization per object. Under the nominal geometry (FDK, SIRT, R$^2$-Gaussian) the reconstructions are blurred and distorted; at $\times4$, NeAT diverges on several objects and Haouchat \etal lose most of the vessel tree of the aneurism, while ours stays closest to the ground truth.


\noindent\textbf{Error magnitude.}
Tab.~\ref{tab:supp_levels} adds the level $\times2$ between the two levels of Tab.~\ref{tab:main}. Ours is nearly flat: from $\times1$ to $\times4$ it changes by 0.21\,dB on the head (39.70 to 39.49\,dB) and 0.85\,dB on the abdomen (36.65 to 35.80\,dB), with 39.64 and 36.07\,dB at $\times2$, whereas Choi \etal lose 6.5 and 4.5\,dB and NeAT already fails at $\times2$ (22.20 and 19.27\,dB). Our geometry errors at $\times2$ (0.005\,px, 0.15\,px, $0.000^\circ$, $0.004^\circ$, $0.47^\circ$) are of the same order as at $\times1$ and $\times4$ (Tab.~\ref{tab:main}).

\noindent\textbf{Random error directions.}
All cases above share one error direction, $s\cdot(2, -2, 0.5, 0.5, 0.5)$. Tab.~\ref{tab:supp_random} instead uses
seven errors with independent components drawn uniformly in $\pm(8\,\mathrm{px}, 8\,\mathrm{px}, 2^\circ, 2^\circ,
2^\circ)$, \ie, up to the $\times4$ magnitude per parameter but with random signs and combinations, on the head and the
abdomen. In residual coordinates, the initial errors reach 37.3\,px in $\delta u$ and 28.9\,px in $\delta v$. In all 14
runs, our final errors were at most 0.010\,px in $\delta u$, 0.59\,px in $\delta v$ and $0.007^\circ$ in $\eta$ and
$\phi$, and $|\theta|$ dropped from $1.27^\circ$ to $0.48^\circ$ (head) and $0.31^\circ$ (abdomen) on average. The
PSNR was $39.44\pm0.38$\,dB on the head and $36.26\pm0.29$\,dB on the abdomen (worst draws 38.67 and 35.67\,dB), on par
with the fixed direction of Tab.~\ref{tab:main} and, even for the worst draw, above every baseline there at $\times1$.
Our results therefore do not depend on the particular error direction of the main experiments.

\begin{table*}[t]
\centering
\caption{Reconstruction on seven further objects (50 views over $360^\circ$, same simulation, error levels and settings as Tab.~\ref{tab:main}): registered 3D PSNR (dB)/SSIM. $^*$A different patient from the abdomen of the main paper. \colorbox{bestbg}{\textbf{Best}} and \colorbox{secondbg}{second best} results are highlighted.}
\label{tab:supp_objects}
\setlength{\tabcolsep}{3pt}
\setlength{\aboverulesep}{0pt}
\setlength{\belowrulesep}{0pt}
\renewcommand{\arraystretch}{1.15}
\footnotesize
\begin{tabular}{l c c c c c c c c}
\toprule
Method & Chest & Foot & Jaw & Abdomen$^*$ & Pelvis & Leg & Aneurism & Mean \\
\midrule
\multicolumn{9}{l}{\textit{Small error ($\times1$)}} \\
\midrule
FDK~\cite{feldkamp1984practical} & 19.84/0.205 & 20.51/0.274 & 22.83/0.314 & 22.86/0.283 & 25.56/0.359 & 23.34/0.314 & 23.27/0.432 & 22.60/0.312 \\
SIRT~\cite{gilbert1972iterative} & 20.69/0.499 & 21.59/0.668 & 23.18/0.377 & 23.72/0.628 & 26.79/0.695 & 24.74/0.843 & 26.17/0.898 & 23.84/0.658 \\
R$^2$-Gaussian~\cite{r2_gaussian} & 21.91/0.648 & 22.40/0.736 & 24.12/0.528 & 24.14/0.729 & 28.11/0.805 & 24.53/0.900 & 26.15/0.932 & 24.48/0.754 \\
\midrule
Panetta \etal~\cite{panetta2008optimization} & 24.19/0.369 & 23.80/0.353 & 27.38/0.492 & 26.65/0.417 & 30.05/0.506 & 29.35/0.357 & 27.75/0.428 & 27.02/0.417 \\
Haouchat \etal~\cite{haouchat2026geometry} & 27.02/0.554 & 25.74/0.476 & 29.00/0.617 & 28.92/0.582 & 32.09/0.659 & 30.78/0.494 & 24.41/0.403 & 28.28/0.541 \\
NeAT~\cite{ruckert2022neat} & 24.06/0.740 & 23.43/0.799 & 27.55/0.600 & 26.43/0.776 & 30.15/0.843 & 35.38/0.967 & 34.96/0.973 & 28.85/0.814 \\
Choi \etal~\cite{choi2026pose} & 30.83/0.909 & 29.56/0.862 & 34.21/0.849 & 32.65/0.925 & 37.15/0.960 & 41.56/0.976 & 38.16/0.985 & 34.87/0.924 \\
Ours (w/o refinement) & \csecond{33.96/0.922} & \csecond{31.55/0.869} & \csecond{35.66/0.872} & \csecond{36.15/0.941} & \csecond{40.51/0.971} & \csecond{48.05/0.992} & \csecond{38.50/0.986} & \csecond{37.77/0.936} \\
Ours & \cbest{34.56/0.928} & \cbest{31.74/0.871} & \cbest{36.06/0.878} & \cbest{36.42/0.943} & \cbest{40.79/0.972} & \cbest{48.24/0.992} & \cbest{39.31/0.987} & \cbest{38.16/0.939} \\
\midrule
\multicolumn{9}{l}{\textit{Large error ($\times4$)}} \\
\midrule
FDK~\cite{feldkamp1984practical} & 15.76/0.158 & 18.22/0.308 & 21.07/0.201 & 18.96/0.207 & 22.00/0.325 & 20.30/0.378 & 22.41/0.392 & 19.82/0.281 \\
SIRT~\cite{gilbert1972iterative} & 16.04/0.317 & 18.82/0.563 & 20.60/0.228 & 18.98/0.385 & 22.11/0.475 & 20.62/0.746 & 25.03/0.872 & 20.31/0.512 \\
R$^2$-Gaussian~\cite{r2_gaussian} & 16.48/0.476 & 19.63/0.619 & 22.17/0.386 & 18.17/0.508 & 21.20/0.608 & 20.44/0.848 & 25.05/0.923 & 20.45/0.624 \\
\midrule
Panetta \etal~\cite{panetta2008optimization} & 22.21/0.367 & 22.31/0.356 & 26.37/0.436 & 24.66/0.419 & 28.51/0.517 & 28.83/0.380 & 27.28/0.442 & 25.74/0.417 \\
Haouchat \etal~\cite{haouchat2026geometry} & 24.40/0.522 & 24.80/0.480 & 22.35/0.352 & 27.06/0.571 & 31.24/0.668 & 27.64/0.399 & 23.52/0.498 & 25.86/0.499 \\
NeAT~\cite{ruckert2022neat} & 15.79/0.339 & 18.44/0.561 & 21.01/0.196 & 17.89/0.417 & 20.16/0.290 & 19.87/0.574 & 22.17/0.596 & 19.33/0.425 \\
Choi \etal~\cite{choi2026pose} & 26.66/0.869 & 25.89/0.816 & 26.94/0.666 & 29.46/0.899 & 33.54/0.946 & 34.84/0.947 & 33.67/0.972 & 30.14/0.874 \\
Ours (w/o refinement) & \csecond{33.15/0.912} & \csecond{31.11/0.861} & \csecond{34.51/0.858} & \csecond{35.02/0.931} & \csecond{38.70/0.962} & \csecond{43.09/0.989} & \csecond{36.81/0.984} & \csecond{36.05/0.928} \\
Ours & \cbest{34.29/0.926} & \cbest{31.77/0.871} & \cbest{35.25/0.872} & \cbest{35.78/0.940} & \cbest{39.69/0.969} & \cbest{43.41/0.990} & \cbest{38.08/0.985} & \cbest{36.89/0.936} \\
\midrule
\textcolor{gray}{R$^2$-Gaussian, true geometry} & \textcolor{gray}{34.53/0.927} & \textcolor{gray}{31.79/0.871} & \textcolor{gray}{36.23/0.879} & \textcolor{gray}{36.51/0.944} & \textcolor{gray}{40.86/0.972} & \textcolor{gray}{48.72/0.992} & \textcolor{gray}{40.24/0.987} & \textcolor{gray}{38.41/0.939} \\
\bottomrule
\end{tabular}
\end{table*}

\begin{table*}[t]
\centering
\caption{Ours under seven random initial errors per object, drawn independently and uniformly in $\pm(8\,\mathrm{px}, 8\,\mathrm{px}, 2^\circ, 2^\circ, 2^\circ)$ for $(u, v, \eta, \phi, \theta)$ (seed 2027; same draws for both objects). Left: initial and final absolute geometry errors; right: registered 3D PSNR (dB)/SSIM. Worst: largest error / lowest PSNR over the seven draws.}
\label{tab:supp_random}
\setlength{\tabcolsep}{3.5pt}
\setlength{\aboverulesep}{0pt}
\setlength{\belowrulesep}{0pt}
\renewcommand{\arraystretch}{1.15}
\footnotesize
\begin{tabular}{ll ccccc ccccc cc}
\toprule
\multirow{2}{*}{Object} & \multirow{2}{*}{Draw} & \multicolumn{5}{c}{Initial error} & \multicolumn{5}{c}{Final error (ours)} & \multicolumn{2}{c}{PSNR/SSIM} \\
\cmidrule(lr){3-7}\cmidrule(lr){8-12}\cmidrule(lr){13-14}
& & $|\Delta u|$ & $|\Delta v|$ & $|\eta|$ & $|\phi|$ & $|\theta|$ & $|\Delta u|$ & $|\Delta v|$ & $|\eta|$ & $|\phi|$ & $|\theta|$ & w/o ref. & Ours \\
\midrule
\multirow{9}{*}{Head} & r1 & 8.026 & 5.11 & 1.670 & 0.010 & 0.21 & 0.002 & 0.19 & 0.000 & 0.006 & 0.24 & 39.03/0.973 & 39.58/0.976 \\
 & r2 & 7.043 & 28.89 & 0.490 & 0.240 & 1.64 & 0.001 & 0.16 & 0.001 & 0.001 & 1.11 & 38.04/0.970 & 38.67/0.973 \\
 & r3 & 37.270 & 27.06 & 1.590 & 1.880 & 1.85 & 0.000 & 0.13 & 0.001 & 0.007 & 0.09 & 38.00/0.967 & 39.72/0.976 \\
 & r4 & 9.393 & 17.19 & 1.270 & 0.350 & 0.87 & 0.001 & 0.08 & 0.001 & 0.000 & 0.19 & 38.78/0.972 & 39.68/0.976 \\
 & r5 & 35.960 & 11.12 & 1.590 & 1.880 & 1.18 & 0.003 & 0.01 & 0.000 & 0.006 & 0.37 & 38.19/0.969 & 39.72/0.976 \\
 & r6 & 24.176 & 23.98 & 0.590 & 1.690 & 1.16 & 0.000 & 0.06 & 0.001 & 0.001 & 0.60 & 38.14/0.969 & 39.22/0.975 \\
 & r7 & 18.894 & 25.77 & 0.890 & 1.440 & 1.97 & 0.002 & 0.01 & 0.000 & 0.004 & 0.78 & 38.44/0.970 & 39.49/0.975 \\
\cmidrule(lr){2-14}
& \textit{Mean} & 20.109 & 19.87 & 1.156 & 1.070 & 1.27 & 0.001 & 0.09 & 0.001 & 0.004 & 0.48 & 38.37/0.970 & 39.44/0.975 \\
& \textit{Worst} & 37.270 & 28.89 & 1.670 & 1.880 & 1.97 & 0.003 & 0.19 & 0.001 & 0.007 & 1.11 & 38.00/0.967 & 38.67/0.973 \\
\midrule
\multirow{9}{*}{Abdomen} & r1 & 8.026 & 5.11 & 1.670 & 0.010 & 0.21 & 0.005 & 0.30 & 0.000 & 0.001 & 0.06 & 36.31/0.951 & 36.55/0.952 \\
 & r2 & 7.043 & 28.89 & 0.490 & 0.240 & 1.64 & 0.010 & 0.01 & 0.000 & 0.003 & 0.35 & 35.70/0.948 & 36.30/0.952 \\
 & r3 & 37.270 & 27.06 & 1.590 & 1.880 & 1.85 & 0.009 & 0.59 & 0.001 & 0.004 & 0.17 & 35.21/0.943 & 36.28/0.952 \\
 & r4 & 9.393 & 17.19 & 1.270 & 0.350 & 0.87 & 0.007 & 0.24 & 0.001 & 0.004 & 0.20 & 36.07/0.949 & 36.25/0.952 \\
 & r5 & 35.960 & 11.12 & 1.590 & 1.880 & 1.18 & 0.008 & 0.05 & 0.000 & 0.002 & 0.26 & 35.54/0.945 & 36.53/0.953 \\
 & r6 & 24.176 & 23.98 & 0.590 & 1.690 & 1.16 & 0.008 & 0.31 & 0.001 & 0.005 & 0.22 & 35.52/0.946 & 36.26/0.952 \\
 & r7 & 18.894 & 25.77 & 0.890 & 1.440 & 1.97 & 0.007 & 0.14 & 0.001 & 0.004 & 0.88 & 35.05/0.944 & 35.67/0.950 \\
\cmidrule(lr){2-14}
& \textit{Mean} & 20.109 & 19.87 & 1.156 & 1.070 & 1.27 & 0.008 & 0.23 & 0.001 & 0.003 & 0.31 & 35.63/0.946 & 36.26/0.952 \\
& \textit{Worst} & 37.270 & 28.89 & 1.670 & 1.880 & 1.97 & 0.010 & 0.59 & 0.001 & 0.005 & 0.88 & 35.05/0.943 & 35.67/0.950 \\
\bottomrule
\end{tabular}
\end{table*}

\end{document}